\documentclass{article}

\PassOptionsToPackage{numbers, compress}{natbib}

 \usepackage[preprint]{neurips_2025}

\usepackage[utf8]{inputenc} %
\usepackage[T1]{fontenc}    %
\usepackage{hyperref}       %
\usepackage{url}            %
\usepackage{booktabs}       %
\usepackage{amsfonts}       %
\usepackage{nicefrac}       %
\usepackage{microtype}      %
\usepackage[dvipsnames]{xcolor}         %

\usepackage{epsfig}
\usepackage{caption}
\usepackage{float}
\usepackage{placeins}
\usepackage{color, colortbl}
\usepackage{stfloats}
\usepackage{enumitem}
\usepackage{tabularx}
\usepackage{xstring}
\usepackage{multirow}
\usepackage{xspace}
\usepackage{subcaption}
\usepackage[hang,flushmargin]{footmisc}
\usepackage{listings}
\usepackage{soul}
\usepackage{wrapfig}
\usepackage[hang,flushmargin]{footmisc} %
\newcommand{\R}[1]{{%
    \textbf{%
        \ifstrequal{#1}{1}{\textcolor{red}{R#1}}{%
        \ifstrequal{#1}{2}{\textcolor{blue}{R#1}}{%
        \ifstrequal{#1}{3}{\textcolor{magenta}{R#1}}{%
        \ifstrequal{#1}{4}{\textcolor{teal}{R#1}}{%
                           \textcolor{cyan}{R#1}%
        }}}}%
    }%
}}

\definecolor{mygrey}{rgb}{0.7, 0.7, 0.7}
\definecolor{mygrey2}{rgb}{0.5, 0.5, 0.5}
\definecolor{mymaroon}{rgb}{0.53, 0.15, 0.34}
\definecolor{mygreen}{rgb}{0.0, 0.6, 0.0}
\definecolor{mygreen}{rgb}{0.0, 0.647, 0.32}
\definecolor{myred1}{hsb}{1, 1., 0.9}
\definecolor{myred2}{hsb}{1, 0.4, 0.9}
\newcommand{\g}[1]{\textcolor{mygrey}{#1}}

\sethlcolor{white}
\newcommand{\methodname}[0]{{\small\fontfamily{txtt}\selectfont \textcolor{RoyalPurple}{\hl{CAViAR}}}\xspace}

\title{CAViAR: Critic-Augmented Video Agentic Reasoning}

\author{%
  \begin{tabular}{@{}c@{\quad}c@{\quad}c@{}}
    \mbox{Sachit Menon\textsuperscript{1,2}\thanks{Work completed while at Google DeepMind.}} &
    \mbox{Ahmet Iscen\textsuperscript{1}} &
    \mbox{Arsha Nagrani\textsuperscript{1}} \\
    \mbox{Tobias Weyand\textsuperscript{1}} &
    \mbox{Carl Vondrick\textsuperscript{2}} &
    \mbox{Cordelia Schmid\textsuperscript{1}}
  \end{tabular}\\[0.3em]
  \textsuperscript{1}Google DeepMind \quad \textsuperscript{2}Columbia University
}

\begin{document}

\maketitle

\enlargethispage{-\baselineskip} %
\vspace{-1.3em}
\begin{abstract}
    Video understanding has seen significant progress in recent years, with models' performance on perception from short clips continuing to rise. Yet, multiple recent benchmarks, such as LVBench, Neptune, and ActivityNet-RTL, show performance wanes for tasks requiring complex reasoning on videos as queries grow more complex and videos grow longer. In this work, we ask: can existing perception capabilities be leveraged to successfully perform more complex video reasoning? In particular, we develop a large language model agent given access to video modules as subagents or tools. Rather than following a fixed procedure to solve queries as in previous work such as Visual Programming, ViperGPT, and MoReVQA, the agent uses the results of each call to a module to determine subsequent steps. Inspired by work in the textual reasoning domain, we introduce a critic to distinguish between instances of successful and unsuccessful sequences from the agent. We show that the combination of our agent and critic achieve strong performance on the previously-mentioned datasets.
\end{abstract}

\section{Introduction}
\label{sec:intro}

\begin{figure*}[h]
    \centering\vspace{-1em}
    \includegraphics[width=0.95\linewidth]{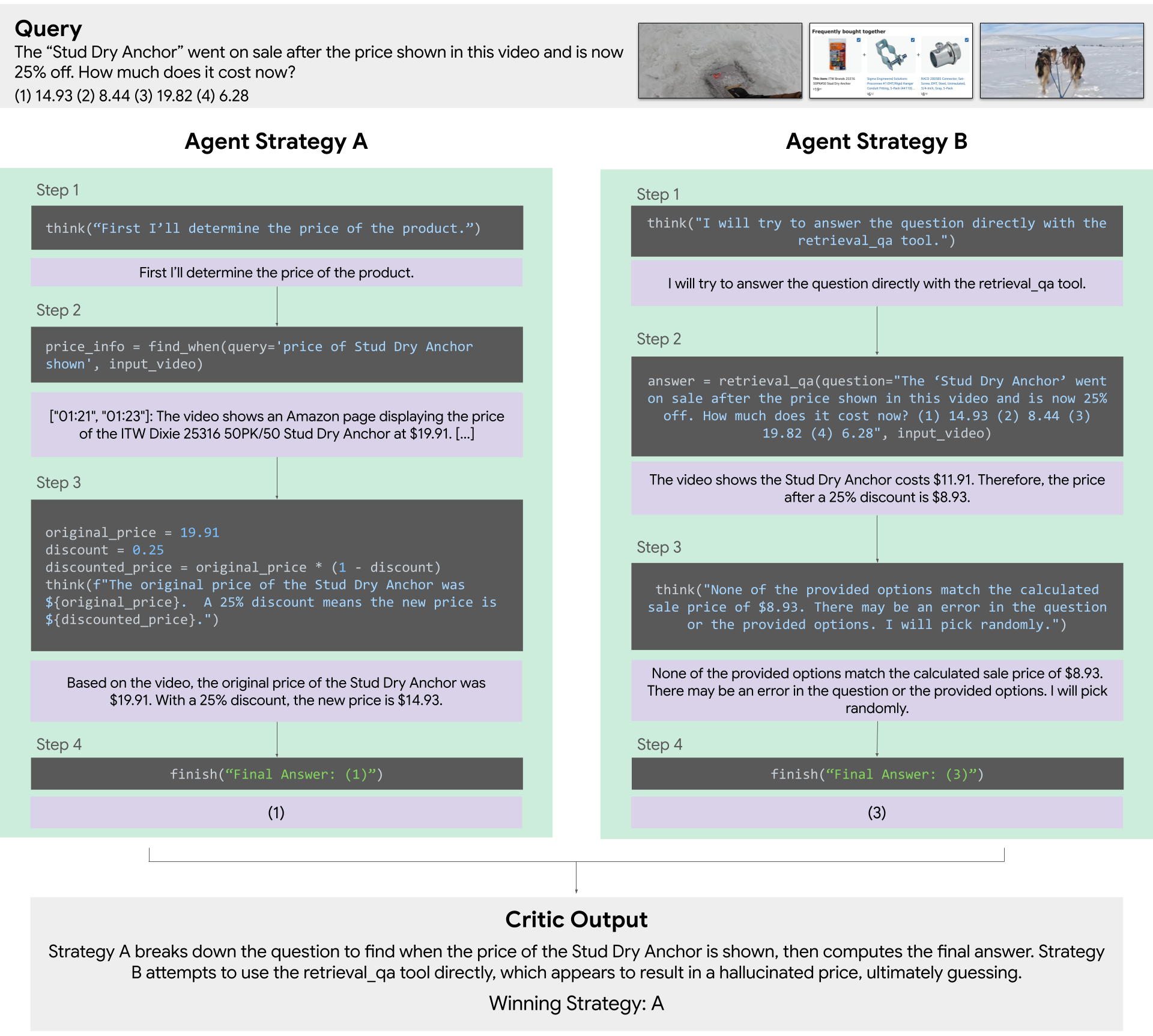}
    \caption{\methodname consists of a reasoning agent that produces sequences of programs to solve video queries with different strategies, followed by a critic that selects the most promising reasoning. Each program invokes visual modules that use the video as input, rather than it being providing as input a single time at the start. We show two strategies here for illustration.}\vspace{-1.5em} %
    \label{fig:example}
\end{figure*}

Advances in video understanding have been propelled by multimodal large‑language models (MLLMs) trained end‑to‑end on visual and text inputs \cite{openai2024gpt4ocard,geminiteam2024geminifamilyhighlycapable,geminiteam2024gemini15unlockingmultimodal,qwen2.5vl}. While these systems have made major strides in basic perception, they often falter with queries demanding compositional, multi‑step reasoning over long videos \cite{wang2024lvbench,nagrani2024neptune,huang2024lita}. Recently, tool-augmented inference has emerged as a powerful class of models towards achieving compositional reasoning in a a variety of domains \cite{zhou2023language,wang2024rethinkingboundsllmreasoning}. These methods  decompose a query into sub‑tasks, invoke specialized modules, and scale to large contexts by selectively ``zooming‑in'' on the relevant portion of the input given appropriate tools. These modular inference approaches offer (i) interpretable decision chains, (ii) graceful scaling as context length and task complexity grow, and (iii) a natural substrate for inference‑time reasoning, as they generate structured reasoning traces by construction.

These approaches have not scaled to the complexity and size of video reasoning tasks in large part because they lack the ability to adapt their procedures during execution. Current tool-based methods typically generate fixed procedures \cite{gupta2023visual,suris2023vipergpt} or use hand-designed stages \cite{min2024morevqa} that are directly executed. Such fixed plans require meticulous prompt engineering of API declarations and in-context examples, as the plan cannot be changed once execution starts. This rigidity requires models to stitch together module outputs that they never actually observe, and systems must decide which tool to use when (and with what inputs) without having seen any cases of where the tools succeed or fail. Consequently, one poor decision snowballs into unrecoverable errors and propagates hallucinations unchecked \cite{tong2024eyes,stanic2024towards}. Video exacerbates this problem because the potential points of failure increase. Rather than rolling out inference plans and hoping for success, if systems adapted their plans to intermediate results as well as explored different strategies, they could find plans that work better depending on the situation. %

We introduce {\large\fontfamily{txtt}\selectfont\textcolor{RoyalPurple}{\hl{C}}}ritic-%
{\large\fontfamily{txtt}\selectfont\textcolor{RoyalPurple}{\hl{A}}}ugmented %
{\large\fontfamily{txtt}\selectfont\textcolor{RoyalPurple}{\hl{Vi}}}deo %
{\large\fontfamily{txtt}\selectfont\textcolor{RoyalPurple}{\hl{A}}}gentic %
{\large\fontfamily{txtt}\selectfont\textcolor{RoyalPurple}{\hl{R}}}easoning (\methodname), which pairs a program‑generating \emph{agent} with a natural‑language \emph{critic}. The agent iteratively composes video modules into sequences of executable programs, considering the result of each before moving on to the next step, enabling adaptability over the course of reasoning. It thereby generates interpretable reasoning traces that lend themselves well to verification, as seen in Agent Strategies A and B in Figure \ref{fig:example}. The critic inspects multiple reasoning traces reflecting different solution strategies and selects the most plausible sequence based on examples it has seen, producing natural language feedback as shown by the Critic Output in Figure \ref{fig:example}. By comparing multiple strategies, the critic enables success in cases where some approaches lead to success while others fail. Together, the agent and critic allow \methodname\ to sidestep brittle tool choices and mitigate hallucinations, achieving state-of-the-art results across multiple tasks and datasets.

\methodname has many benefits over previous modular approaches: it is  interpretable, as each step corresponds to a short program that can be easily examined; it enables scaling the performance of a single underlying model with no additional training; it avoids issues of customizing module selection to a particular domain, as the critic selects the most promising strategy; it affords compositionality without extreme hand-tuning of module definitions, as the reasoning agent can see module outputs in-context to decide how to use them; and it is general, easily incorporating any additional video modules or Python code that may be useful for a particular domain.

In summary, our contributions are:
\begin{enumerate}[leftmargin=*, labelindent=0pt]
    \item We present a framework for video understanding using agentic reasoning on video modules.
    \item We introduce a critic, which highlights how small amounts of feedback can be used towards substantial performance improvements while avoiding dataset-specific tuning of modules. 
    \item We show \methodname demonstrates strong performance on multiple tasks such as temporal localization with reasoning and long video question answering on recent, difficult datasets.
\end{enumerate}

\section{Related Work}
\label{sec:related}

\subsection{LLMs, Augmented Inference Procedures, and Text Reasoning}

A wide range of work has emerged in recent years using large language models to better solve text reasoning tasks. 
\citet{cobbe2021trainingverifierssolvemath} introduce process supervision, the idea of using information from the model's stated chain-of-thought reasoning to select better answers. \citet{zelikman2022star} filter for chain-of-thought rationales with correct final answers and train the base model further on those. Noting that a correct final answer may not correspond to correct reasoning, \citet{hosseini2024v} build on this by incorporating a verifier trained on pairs of correct and incorrect solutions to predict a scalar value indicating whether a given candidate rationale is correct, using it at inference time to select the best answers. \citet{zhang2024generative} observe that training to produce a scalar value may miss out on the benefits of pretrained LLMs' inherent reasoning capabilities. They point out that the scalar-output verifier approach cannot make use of human-written natural language critiques for why a given candidate answer was wrong rather than just whether it was correct or not, showing that using natural language generation with such critiques to rank reasoning outputs outperforms discriminative scalar-output verifiers. ToRA empowers LLMs with tools for mathematics such as computation libraries callable from Python (e.g., \texttt{sympy}), sequentially using code to call tools for math problems. The tool call trajectories are trained to imitate a stronger model, then are filtered for correctness to train the base agent further. These works consider the domain of math word problems, due to its relative ease of verification and the natural breakdown of problems into a structured form of steps and their results. 

\subsection{Video Understanding and Video Reasoning}

Substantial recent progress in video understanding and video reasoning has come from multimodal language models \cite{openai2024gpt4ocard,geminiteam2024geminifamilyhighlycapable,geminiteam2024gemini15unlockingmultimodal,qwen2.5vl}. These models are trained with a mixture of video and text inputs to perform next-token prediction of text tokens. SeViLA finetunes the BLIP-2 text-image model to perform localization and QA tasks for video \cite{yu2023self}. \citet{zhang2024video} demonstrate that synthetic data to copy a stronger model can be a powerful signal for weaker models to understand video. LITA \cite{huang2024lita} finds that despite performance on temporal localization benchmarks, video models struggle to perform localization tasks that requires reasoning. They introduce the ActivityNet-RTL dataset to evaluate this reasoning temporal localization, and show that training with synthetic data from strong teacher models substantially improves performance on this task.

\subsection{Agents and Tool Use in Vision}

As the ability of large language models to use tools and perform agentic reasoning has grown, some work has emerged in the visual arena making use of these abilities. Visual Programming \cite{gupta2023visual} and ViperGPT \cite{suris2023vipergpt} prompt a language model to produce a program using computer vision tools that solves input queries, philosophically following \citet{johnson2017inferring}, which aimed to perform visual reasoning by learning to generate programs prior to the advent of large language models. Visual Programming relies on many hand-written examples of programs using the given tools. ViperGPT instead defines an API for the provided tools and uses fewer examples, as well as showing some results in the video domain. Further work has found that both the choice of modules per dataset \cite{khandelwal2023analyzing} and the examples constructed \cite{stanic2024towards} substantially influence the performance of single-program approaches. AVIS \cite{hu2023avis} goes beyond single-program approaches to use tools with tree search for knowledge-intensive image question answering. They define a transition graph of valid tool sequences, which define the tools a language model can choose to use next after each tool. The approach successfully goes beyond fixed procedures, but relies heavily on human knowledge and examples as well as being focused on single images.

Various methods have also emerged recently in this direction specific to video. LLoVi \cite{zhang2023simple} uses a frozen captioner on frames and passes them to an LLM to produce an answer. VideoAgent \cite{wang2024videoagent} and VideoTree \cite{wang2024videotree} both use modified inference procedures for video primarily in their selection of frames, both using captions and text similarity. VideoAgent captions every frame and uses an additional frozen image-text similarity model (CLIP \cite{radford_learning_2021}) to choose frames, iteratively using the captions from those frames with an LLM to produce an answer. They repeat this caption-and-retrieve process, prompting an LLM to ask whether the question has been answered confidently each time until a confidence threshold is reached. VideoTree uses a tree-based representation to build a set of key frames to caption via clustering, then using these captions with an LLM to produce a final answer. MoReVQA \cite{min2024morevqa} identifies shortcomings in the single-program approach when applied to video, in particular that such planning methods are brittle, failing when inputs do not conform to the expectations set out in the generated program. They define a three-stage procedure specific to video using visual modules: event parsing, grounding, and reasoning, leading to a final prediction. These fixed stages are used to produce predictions for video question answering queries given few-shot examples of how to use the tools provided.

\begin{wrapfigure}{r}{0.625\columnwidth}
    \vspace{-3em}
    \includegraphics[width=0.625\columnwidth]{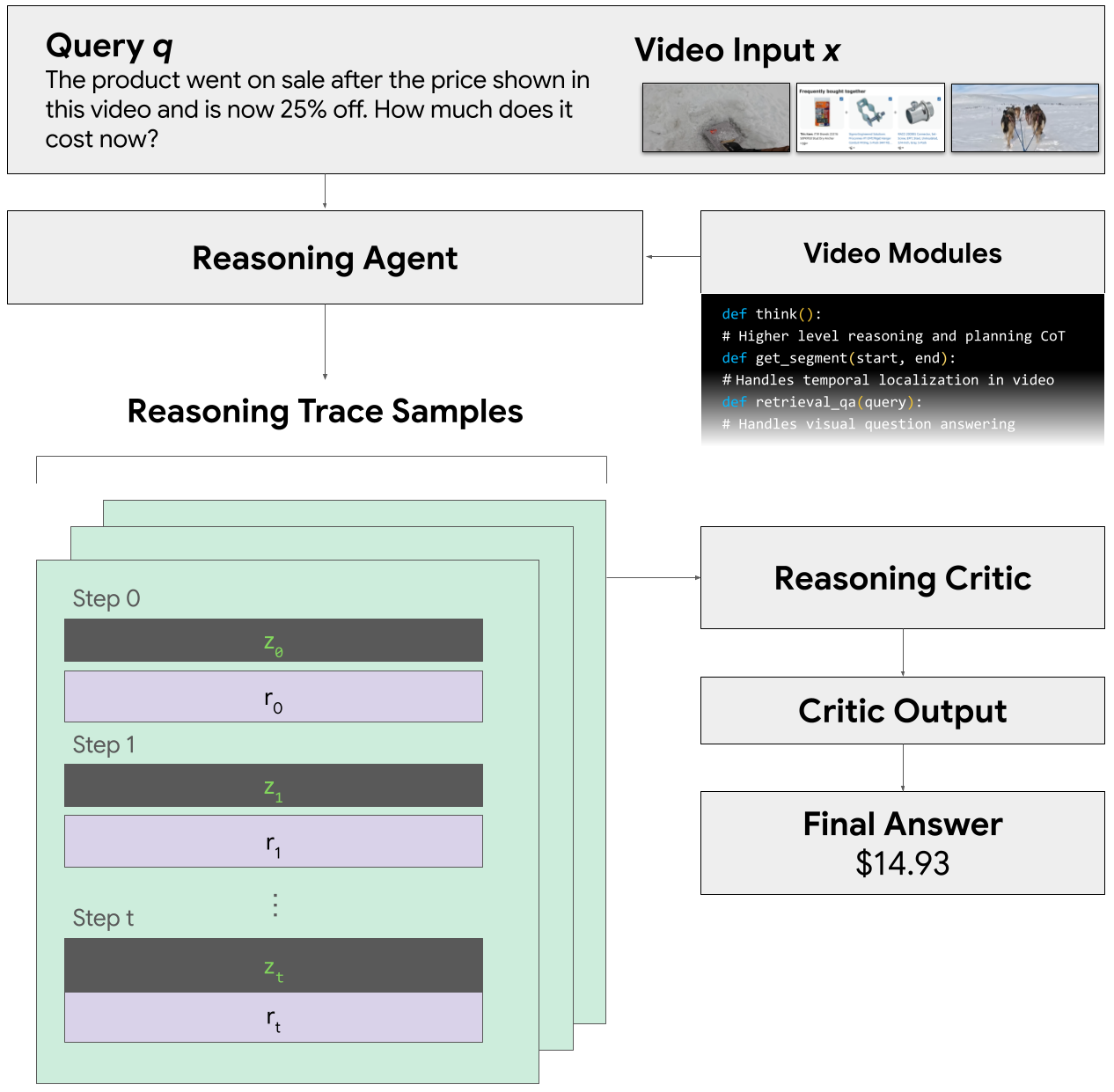}
    \captionsetup{width=0.62\columnwidth}
    \caption{\textbf{The \methodname system.} The reasoning agent generates reasoning traces to solve the query using video modules. The critic selects the best, yielding a final answer.}
    \vspace{-1.5em}
    \label{fig:system}
\end{wrapfigure}

\section{Method}
\label{sec:method}
\subsection{Overview}

Given a video and a prompt, the reasoning agent iteratively generates and executes programs, using provided video-processing modules and a Python interpreter to ultimately converge to a final answer.
We refer to the resulting sequences as reasoning traces or trajectories, as they comprise the full history of steps taken by the reasoning agent to understand the input, reason through the question, and produce a final answer. The reasoning traces produced by the agent are provided to the critic, which provides natural language feedback on their likelihood of success. The feedback from the critic is used to select one of the candidate reasoning traces and its associated final answer. Figure \ref{fig:system} shows the overall procedure. We detail this process in this section.

\subsection{Reasoning Agent}

Given a visual (or multimodal) input $x$ and a textual query $q$, \methodname first generates a program $z_1 = \pi(q)$ with the reasoning agent. The reasoning agent $\pi$ is by design fairly simple. We provide it an API for the provided video modules (discussed in the next section) in the form of Python function headers and docstrings.

We note that the exact specifications of outputs need not be as exact as when using a single-program system, as discussed later in this section. The input to the reasoning agent $\pi$, this API (found in full in the Supplementary Material) is comprised of similar definitions for every video module.

The execution engine $\phi$ executes this program on the inputs using the code interpreter and the provided video modules, obtaining the result $r_1 = \phi(x,z_1)$. This already can be a competitive approach in some circumstances, as demonstrated by \cite{suris2023vipergpt,gupta2023visual}. Yet, the single-program approach has multiple limitations that make its application difficult in practice, as we show in our ablations. It requires careful design of the modules provided, their API descriptions, and the associated examples to ensure the program generator can write programs that use their outputs without being able to see them and decide how to use them next \cite{stanic2024towards}. Unlike previous work \cite{suris2023vipergpt,gupta2023visual}, we explicitly \textit{do not} provide any hand-designed examples of programs or reasoning traces to the agent. While this would likely improve performance, annotating full programs or reasoning traces is challenging and requires expert annotators, which would make the method less usable in practice.

The procedure does not end with the first generated program and result. Rather, the reasoning agent is given both the generated program and its result and asked to produce another program as the next step towards a solution, $z_2 = \pi(q, z_1, r_1)$. This procedure proceeds until the reasoning agent determines it has found a solution, producing a new program $z_i$ based on the full history of program and results and executing it to obtain the next result. 

The reasoning agent thus produces a sequence $S=(z_1, r_1, z_2, r_2, \ldots, z_n, r_n)$ culminating in the final answer in $r_n$. (Note that the number of steps, $n$, is not fixed and can be freely decided by the reasoning agent.) We refer to this sequence $S$ as a reasoning trace, or equivalently a trajectory.

\subsection{Video Modules and Their API}\label{sec:modules}

We aim for a minimal, yet general set of video modules for the experiments we present here. The framework allows for any additional modules to be added on. In all cases, the full range of Python as a programming language is already taken as given -- these modules are what we provide on top, thus we do not explicitly note e.g., a `calculator' tool here. We provide an overview of the modules used and our reasons for including them here, with further implementation details in the Supplementary Material. As a note of terminology, we follow the broader multi-agent literature and refer to the base model being used with different prompts or inference schemes as agents (or subagents) while other operations, such as programmatic ones, are referred to as tools.

\textbf{Visual Retrieval + QA} (\texttt{\textbf{retrieval\_qa}}). One of the most fundamental capabilities needed to understand long videos is the ability to obtain the most visually relevant frames to a given query and use them to get the information corresponding to said query. Retrieval of relevant frames also allows for further intermediate interpretability via visual inspection. In order to use a single model, we perform this retrieval by prompting the model with a sliding window of the frames up to the limit of its context window. At a high level, it uses the visual capabilities of the underlying model to directly ask which individual frames are visually relevant in the given window; the identified frames are then considered to respond to the query. Please see Supplementary Material for further discussion.

\textbf{Temporal Grounding (\texttt{\textbf{get\_segment}})}. Aside from considering the relevant portion of a video based on its visual content, another important ability is to ground temporal information to the video -- that is, select the most relevant part of the video based on explicit times. Given a start and end timestamp (and the frame rate for any input video as given), this tool allows the agent to use temporal information in this way by simply trimming out the relevant segment. This can also enable better interpretability in knowing which part of the video the agent chose to consider.

\textbf{Temporal Localization (\texttt{\textbf{find\_when}})}. The natural converse, then, is to identify time information from the video. Given a query corresponding to an event or action, this subagent aims to determine potential ranges of time (identified by a start timestamp and an end timestamp) that may correspond to the given query along with a brief description of what led it to output each range. 
If the video is longer than the context length available, this subagent also employs a windowed approach, considering the information corresponding to each window and returning any possible ranges found from each window. Its specialized instructions guide it to prioritize recall over precision so as to give the reasoning agent as much information as possible.

\textbf{ASR Understanding} (\texttt{\textbf{asr\_understanding}}). When a video contains speech, it often contains critical context to the visual content of the video. This subagent takes a query and attempts to identify any relevant information from an automatic transcription of the speech in the video. If the transcript is too long for the context, it obtains information from each piece up to its context window, then tries to consolidate the information obtained from the different parts of the transcript.

\textbf{Think} (\texttt{\textbf{think}}). This basic tool simply allows the reasoning agent to perform a step of explicit verbal reasoning to plan its next tool use before proceeding. It allows the reasoning agent to use a step to reason instead of to use a tool and returns the text reasoning from the agent on how to proceed. 

\textbf{Completion} (\texttt{\textbf{finish}}). The agent indicates its final answer and ends the agentic inference procedure.

\subsection{Reasoning Critic}

Simply taking the final answer from a single sequence can be a competitive approach, as we later show. Yet, especially when multiple available modules seem like they could lead to a solution, performance can degrade with the single-sequence approach, as we show in our ablations. We note the reasoning agent can make two major categories of errors: faulty reasoning, or tool malfunction. %

The reasoning critic, $c$, presents a natural solution to this problem. Rather than sampling a single sequence and accepting the result, we can produce multiple $S_i$, each representing different strategies the reasoning agent can use to obtain an answer. The reasoning critic then critiques the sampled strategies according to its inherent reasoning capabilities and any examples it has been provided to learn from. Attempting to fact-check from the visual input would raise a chicken-and-egg problem: the critic would in some regards need more reliable visual processing capabilities than the agent itself to make corrections rather than fall for the same hallucinations.
In addition, language models have been shown to be poor evaluators of their own predictions without additional information or comparison \cite{kamoi2024can,huang2023large}; we demonstrate in Section \ref{sec:selfeval} that such self-evaluation results in reduced accuracy.
Instead, the critic asks: based on the examples of reasoning traces that have succeeded and failed, how likely is this reasoning trace to succeed? That is, how successful are the reasoning and tool calls given what has worked in the past? Considering these text reasoning traces are also less intensive than entire videos, the feedback from the critic can be used to pick the most promising strategy, as we primarily consider here, or could be used in other ways, such as to provide feedback for further steps from the reasoning agent. %

Given reasoning traces $S_i$ corresponding to various sampled strategies, the reasoning critic $c$ aims to provide a critique in the form of natural language feedback for the given strategies; its ultimate goal is to provide a recommendation for which strategy may be the most promising. 
To obtain trajectories corresponding to markedly different strategies, we provide the reasoning agent different subsets of modules that can lead to a final answer, observing that standard temperature sampling did not produce significant variation in outputs in the video modeling setting.
If a module can directly produce an answer of the appropriate type for the query, it is directly applied with guidance to write out its reasoning to provide the critic more information. (For instance: the \texttt{retrieval\_qa} module produces valid outputs for QA problems while the temporal grounding tool does not.)

Inspired by work in the RLHF space, we note that it is easier both for models and humans to identify a preference between presented options than to assign a well-calibrated numeric score to each independently \cite{christiano2017deep}. We therefore present all sampled strategies to the reasoning critic at once, prompting it to critique the given strategies and identify any that could be considered `winning strategies.'
We demonstrate that the critic can achieve strong performance using a small number of in-context examples. Each in-context example is constructed with a question, each sampled strategy, an optional brief critique, and a list of the winning strategies.

\section{Evaluation}
\label{sec:evaluation}

\methodname can handle a variety of tasks involving multimodal inputs depending on the tools provided. We showcase this with two tasks across three different datasets: two multiple-choice complex long video question answering settings on the LVBench \cite{wang2024lvbench} and Neptune \cite{nagrani2024neptune} datasets, and the reasoning temporal localization task on the ActivityNet-RTL dataset (\cite{huang2024lita}).%
(Additional comparison to prior work on the EgoSchema dataset \cite{mangalam2023egoschema} is presented in the Supplemental Material.)

For our primary results, we consider Gemini Flash 1.5 with 32k token (roughly 120 frames) context as our base model due to a combination of 1) its base ability to process videos 2) its ability to act as an agent given instructions 3) cost, speed, and compute/credit availability considerations. Towards the goal of scaling performance of one model, we implement the critic with the same base model as the agent.
To show the generality of our method, we also show some results with GPT-4o-mini in the Supplementary Material. %
Selecting subsets that result in different strategies leads to $3$ strategies for the tasks we consider given the modules described in Section \ref{sec:modules}. We use $4$ in-context examples for the critic for each dataset.
Full details can be found in the Supplementary Material.

\subsection{Complex Video QA}

\begin{table}[t!]
\centering
\begin{minipage}{.48\textwidth}
    \caption{\textbf{LVBench Results}. We report accuracy on the evaluation set. Direct inference and \methodname results use Gemini 1.5 Flash. Other results reported from dataset leaderboard \cite{wang2024lvbench}.}
    \centering
         \resizebox{\columnwidth}{!}{\begin{tabular}{c l c}
        \toprule
        & &\textbf{Accuracy (\%)} $\uparrow$ \\
        \midrule
        & Kangaroo \cite{liu2024kangaroo} & 38.3 \\
        & GLM-4V-Plus-0111 \cite{wang2023cogvlm} & 48.7 \\
        & Qwen2.5-VL (7B) \cite{qwen2.5vl} & 45.3 \\
        & Qwen2.5-VL (72B) \cite{qwen2.5vl} & 47.3 \\
        & mPLUG-Owl3 \cite{ye2024mplugowl3longimagesequenceunderstanding} & 43.5 \\
        \midrule
        & Direct Inference & 46.0 \\
        & \methodname & \textbf{62.0} \\
        \bottomrule
    \end{tabular}}
    \label{tab:lvbench}
\end{minipage}%
\hfill
\begin{minipage}{.48\textwidth}
    \caption{\textbf{Neptune Results}. We report accuracy on the evaluation set. Direct inference and \methodname results use Gemini 1.5 Flash. Other results reported from original dataset \cite{nagrani2024neptune}.}
    \centering
         \resizebox{\columnwidth}{!}{\begin{tabular}{c l c}
        \toprule
        &  & \textbf{Accuracy (\%)} $\uparrow$ \\
        \midrule
         & VideoLLaMA2 \cite{cheng2024videollama} & 44.7 \\
         & VideoLLaMA2 with ASR \cite{cheng2024videollama} & 49.3 \\
        & LLaVA-OneVision \cite{li2024llava} & 66.2 \\
        & InternVL2-8B \cite{chen2024internvl2} & 57.1 \\
        & MiniCPM-v \cite{yao2024minicpm} & 56.6 \\
        \midrule
        & Direct Inference & 51.4  \\
        & Direct Inference with ASR & 74.9 \\
        & \methodname & \textbf{77.2}  \\
        \bottomrule
    \end{tabular}}
    \label{tab:neptune}
\end{minipage}
\end{table}

First, we explore video question answering. With the rise of powerful video models, difficult benchmarks such as LVBench \cite{wang2024lvbench} and Neptune \cite{nagrani2024neptune} have emerged, each with unique challenges. 

We consider the LVBench dataset \cite{wang2024lvbench} (CC-BY-NC-SA-4.0 license) to explore the performance of our method on visually challenging, very long videos. LVBench focuses on visual understanding of particularly long videos, spanning multiple hours. The dataset covers a wide range of domains, including everyday activities, sports, entertainment, and more. Audio and speech information are not allowed, making it a challenging visual-only measure of performance. No supervised methods are available to compare to as there is no training set. Neptune \cite{nagrani2024neptune} (CC-BY/Apache 2.0 license) uses audio information such as speech alongside long videos consisting of many frames. It also covers a wide range of domains, with a focus on varied, complex types of question, such as counting and temporal ordering of events. We evaluate on the full dataset (Neptune-Full) rather than the MMH

Results for LVBench can be seen in Table \ref{tab:lvbench}. \methodname yields a $13\%$ absolute improvement over the state of the art. Results for Neptune can be seen in Table \ref{tab:neptune}. \methodname yields meaningful improvement even over direct inference with ASR provided manually, showing the agentic approach not only successfully integrates the ASR information with the \texttt{asr\_understanding} module but goes further. 

\subsection{Reasoning Temporal Localization}

\begin{figure}[b!]
    \includegraphics[width=\columnwidth]{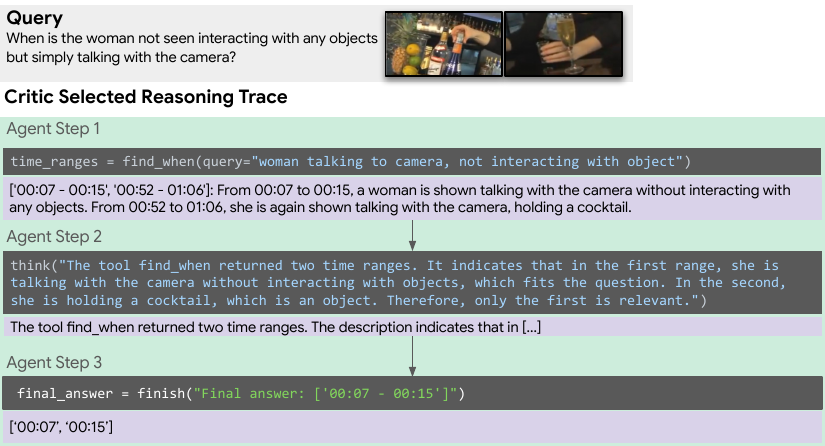}
    \caption{\textbf{Reasoning temporal localization.} Queries require identifying a time range for events/actions that requires reasoning. Here, the critic chooses a strategy which initially finds multiple ranges but correctly reasons which should be included in a final answer. 
    }
    \label{fig:rtl_ex}
\end{figure}

The next task we consider is reasoning temporal localization (RTL). Traditional temporal localization aims to identify a time range in a video that corresponds to when a directly visible action or event occurred. 
\citet{huang2024lita} introduce the reasoning time localization task and the ActivityNet-RTL dataset (videos MIT license) with the goal of evaluating queries that require reasoning on top of the localization capabilities to determine an answer. For instance, consider Figure \ref{fig:rtl_ex}. Asking to localize when the woman is communicating with the camera but not interacting with any objects requires not only the direct perception capability to recognize when she is talking to the camera, but also the reasoning capabilities to understand negation and pick the particular instance when she is not interacting with any objects. We include this task to showcase \methodname's capabilities beyond standard question answering.

\begin{wraptable}{r}{0.45\columnwidth}
    \vspace{-1.3em}
    \caption{\textbf{ActivityNet-RTL Results}. We report mIOU on the evaluation set. Direct inference results use Gemini 1.5 Flash. Other results reported from \cite{huang2024lita}.
    }
    
    \centering
         \resizebox{0.375\columnwidth}{!}{
         \begin{tabular}{c l c}
        \toprule
        & &\textbf{mIOU} $\uparrow$ \\   
        \midrule
        & \g{LlaVa-Finetuned \cite{liu2023visual}} & \g{14.6} \\
        & \g{SlowFast-LlaVa \cite{huang2024lita}} & \g{17.5} \\
        & \g{LITA \cite{huang2024lita}} & \g{28.6}\\
        \midrule
        & Direct Inference & 23.0 \\
        & \methodname & 32.3 \\
        \bottomrule
    \end{tabular}
    }
    \vspace{-3.5em}
    \label{tab:rtl_results}
\end{wraptable}

We compare to the fully supervised state-of-the-art LITA as well as to the base MLLM.
The metric for RTL is mean intersection-over-union (mIOU), the average overlap with the ground truth range over the total union of both across all data points. Table \ref{tab:rtl_results} shows the results. \methodname improves on direct usage of the same model by more than $9$ points.

\section{Ablations and Qualitative Results}

We perform ablations to elucidate what makes \methodname work with the LVBench and Neptune datasets.

\begin{figure}[h!]
    \includegraphics[width=\columnwidth]{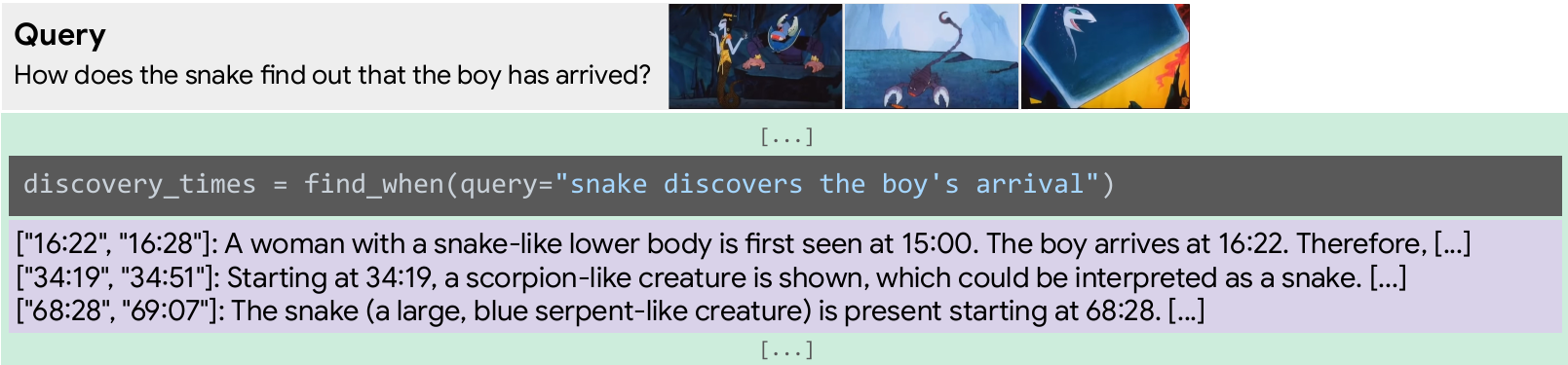}
    \caption{\textbf{Module reliability.} The \texttt{find\_when} module often reports substantial amounts of tangentially relevant, distracting information when applied to long videos, often leading the agent with all modules to failure.
    }
    \label{fig:findlv}
\end{figure}

\textbf{Critic vs no critic} One may think on first glance that the reasoning agent armed with video modules could be sufficient on its own. Table \ref{tab:ablations} shows that not using the critic results in a stark drop in performance, almost to the level of direct inference. Why is this? Looking into the error cases, we find that the \texttt{find\_when} module appears to be less reliable for very long videos, but given all modules, the agent still tries to use it. This effect is particularly pronounced for LVBench, which has very long videos. See Figure \ref{fig:findlv} for an example. In this example, the agent tries to use this module to find when the snake discovers the boy, but in its attempt to yield all potentially relevant information, it reports 21 time ranges. For instance, one centers around the boy being found by a scorpion. This module failure ultimately harms the performance of the agent given all modules. The critic sees options sampled with multiple subsets of modules, and is able to pick more accurate strategies than the one the agent picks by default. Prior work analyzing programmatic approaches \cite{khandelwal2023analyzing} identified their performance is similarly dependent on the set of modules specified per dataset.

\begin{figure}[h!]
    \centering
    \includegraphics[width=0.9\columnwidth]{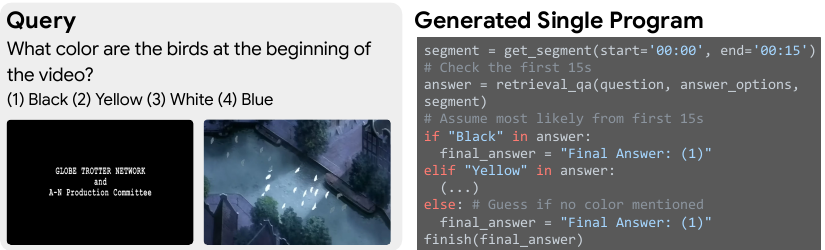}
    \caption{
\textbf{Modular assumptions hinder single program performance.} In this case, the program assumes the answer appears within the first 15 seconds and that \texttt{retrieval\_qa} output includes it exactly. Finding no birds in that segment, it lacks viable answers and resorts to a guess. 
    }
    \label{fig:singleprogram}
\end{figure}

\textbf{Reasoning agent vs single program}
How does generating a single program rather than using the reasoning agent to produce a reasoning trace perform? We present the base model the same API for the modules given to the reasoning agent and instead ask it to generate a single program that solves the query, like in \cite{suris2023vipergpt} or \cite{gupta2023visual}. As seen in Table \ref{tab:ablations}, using a single program results in nearly random performance in this setting. We find this is due to an overwhelming number of guesses or exceptions from incorrect assumptions of modules when the model tries to write a single program; unlike the agent, a single-program approach is not able to inspect the outputs of a module to see what they actually are like beyond the initial impression from the API. See Figure \ref{fig:singleprogram} for a representative example. In this example, the program assumes the answer can be found in the first 15 seconds and that the \texttt{retrieval\_qa} output will contain an exact answer. As the first 15 seconds do not contain the relevant information, this fails, and the program resorts to a guess. This follows observations from \citet{stanic2024towards}, who illustrate that single-program approaches are dependent on extensive hand-tuning of API descriptions and expert-annotated examples of successful programs using the provided modules.

\begin{table}[t!]
\centering
\begin{minipage}{.48\textwidth}
\caption{\textbf{Ablations.}. Items in italics require oracle validation accuracy and are shown only to better understand the method.}
\centering
\resizebox{\columnwidth}{!}{
\begin{tabular}{l c c}
\toprule
& \multicolumn{2}{c}{\textbf{Accuracy (\%)} $\uparrow$} \\
\cmidrule(lr){2-3}
& LVBench & Neptune \\
\midrule
Single Program  & 27.1 & 28.7 \\
\textit{Single Program (Optimal Fixed Modules)}  & 43.3 & 42.0 \\
Agent  & 47.1 & 72.5 \\
\textit{Agent (Optimal Fixed Modules)} & 59.8 & 76.5 \\
Agent + Critic & \textbf{62.0} & \textbf{77.2} \\
\bottomrule
\end{tabular}
}
\label{tab:ablations}
\end{minipage}%
\hfill
\begin{minipage}{.48\textwidth}
\vspace{-.3em}
\caption{\textbf{Self-Evaluation vs Critic.} Using self-evaluation \cite{wang2024videoagent} to terminate based on a confidence score \textit{reduces} performance, while the critic substantially increases it.}
\centering
\resizebox{\columnwidth}{!}{
\begin{tabular}{c l c}
\toprule
& &\textbf{Accuracy (\%)} $\uparrow$ \\
\midrule
& Agent & 47.1 \\
& Agent + Self-Eval Module \cite{wang2024videoagent} & 39.9 \\
& Agent + Critic & \textbf{62.0} \\
\bottomrule
\end{tabular}
}
\label{tab:selfeval}
\end{minipage}
\vspace{-1em}
\end{table}

\textbf{Module configuration.} How much of the critic's performance can be attributed to finding a configuration of modules that works in general rather than per-query? To compare this, we find the optimal fixed subset of modules possible, running each valid combination and considering the evaluation accuracy across the dataset. (Note that rather than oracle per-question accuracy with variable modules, this uses oracle full-dataset accuracy with fixed modules.)  
For LVBench, it uses both \texttt{get\_segment} and \texttt{retrieval\_qa}; for Neptune, \texttt{get\_segment}, \texttt{retrieval\_qa}, and \texttt{asr\_understanding}. 
Note that this type of module selection typically requires oracle validation, and is therefore not a realistic option in most settings; we present it here to better understand \methodname. Using the optimal subset of modules, the agent performance improves substantially, yet the critic goes even further.

\textbf{Confidence self-evaluation vs critic.} \label{sec:selfeval} \citet{wang2024videoagent} introduce a `confidence' module for deciding when to lock in a final answer. Rather than the agent choosing when to finish and report a final answer on its own as in our work, after each step they prompt an LLM and ask: on a scale of 1-3, how confident are you in this answer? The system repeatedly tries to gather new information until this `self-evaluation' module reports a confidence of 3, at which point the answer is reported. This could be considered an alternative in some ways to our critic; rather than producing a critique given different strategies based on examples as our critic does, it critiques a single strategy based on the LLM's prior knowledge and ability to self-evaluate. We consider this comparison on LVBench in Table \ref{tab:selfeval}. We find that in fact, using self-evaluation leads to a substantial \textit{drop} relative to the agent deciding when to produce an answer on its own. We find this stems from poor calibration of the self-evaluation scores, for instance giving low confidence scores until repeatedly gathering irrelevant information. This follows work showing LLMs are not good judges of their own reasoning in isolation \cite{kamoi2024can,huang2023large}.

\section{Conclusion}
\label{sec:conclusion}

In this work, we present \methodname, a framework for video understanding with agentic reasoning on video modules augmented by reasoning critics. It uses an LLM-driven agent to write sequences of programs, executed using visual modules. The critic selects the most promising strategies, which we show allows for flexible use of modules without hand-customized module selection per dataset. \methodname thus uses additional compute with the same base model to improve performance. This could be a limitation in compute-limited settings or when a strong instruction-following base MLLM is not available. The broader impacts will result in improvements for positive applications such as enhancing accessibility, but could also include e.g., unwanted surveillance. Overall, we show \methodname achieves strong performance on multiple reasoning-intensive video tasks.

\newpage
\section*{NeurIPS Paper Checklist}

\begin{enumerate}

\item {\bf Claims}
    \item[] Question: Do the main claims made in the abstract and introduction accurately reflect the paper's contributions and scope?
    \item[] Answer: \answerYes{} %
    \item[] Justification: The paper claims to advance the state-of-the-art in video reasoning by addressing key issues limiting modular, tool-based approaches using an agent and a critic. The numbers correspond to these state-of-the-art claims, and ablations show the agent and critic are critical to these results.

\item {\bf Limitations}
    \item[] Question: Does the paper discuss the limitations of the work performed by the authors?
    \item[] Answer: \answerYes{} %
    \item[] Justification: The paper touches on some limitations in the Conclusions.

\item {\bf Theory assumptions and proofs}
    \item[] Question: For each theoretical result, does the paper provide the full set of assumptions and a complete (and correct) proof?
    \item[] Answer: \answerNA{} %
    \item[] Justification: NA

    \item {\bf Experimental result reproducibility}
    \item[] Question: Does the paper fully disclose all the information needed to reproduce the main experimental results of the paper to the extent that it affects the main claims and/or conclusions of the paper (regardless of whether the code and data are provided or not)?
    \item[] Answer: \answerYes{} %
    \item[] Justification: All models, settings, prompts, and agentic architecture details are provided in the paper and the appendix in full.

\item {\bf Open access to data and code}
    \item[] Question: Does the paper provide open access to the data and code, with sufficient instructions to faithfully reproduce the main experimental results, as described in supplemental material?
    \item[] Answer: \answerNo{} %
    \item[] Justification: \answerTODO{} All experiments use publicly-available data. Code is not available at the time of submission.

\item {\bf Experimental setting/details}
    \item[] Question: Does the paper specify all the training and test details (e.g., data splits, hyperparameters, how they were chosen, type of optimizer, etc.) necessary to understand the results?
    \item[] Answer: \answerYes{} %
    \item[] Justification: All datasets are pointed out to only use the evaluation sets. All sampling parameters and exact prompts are provided in the appendix.

\item {\bf Experiment statistical significance}
    \item[] Question: Does the paper report error bars suitably and correctly defined or other appropriate information about the statistical significance of the experiments?
    \item[] Answer: \answerNo{} %
    \item[] Justification: Experiments are run in a deterministic setting.

\item {\bf Experiments compute resources}
    \item[] Question: For each experiment, does the paper provide sufficient information on the computer resources (type of compute workers, memory, time of execution) needed to reproduce the experiments?
    \item[] Answer: \answerYes{} %
    \item[] Justification: All experiments use commercial APIs to query models.
    
\item {\bf Code of ethics}
    \item[] Question: Does the research conducted in the paper conform, in every respect, with the NeurIPS Code of Ethics \url{https://neurips.cc/public/EthicsGuidelines}?
    \item[] Answer: \answerYes{} %
    \item[] Justification: The research in the paper does not have substantial direct potential for harm or safety concerns as described in the NeurIPS Code of Ethics beyond any work advancing general video reasoning.

\item {\bf Broader impacts}
    \item[] Question: Does the paper discuss both potential positive societal impacts and negative societal impacts of the work performed?
    \item[] Answer: \answerYes{} %
    \item[] Justification: Potential broader impacts of the work are touched on in the Conclusions.
    
\item {\bf Safeguards}
    \item[] Question: Does the paper describe safeguards that have been put in place for responsible release of data or models that have a high risk for misuse (e.g., pretrained language models, image generators, or scraped datasets)?
    \item[] Answer: \answerNA{} %
    \item[] Justification: NA

\item {\bf Licenses for existing assets}
    \item[] Question: Are the creators or original owners of assets (e.g., code, data, models), used in the paper, properly credited and are the license and terms of use explicitly mentioned and properly respected?
    \item[] Answer: \answerYes{} %
    \item[] Justification: License information for datasets used are noted. Models used are commercial.

\item {\bf New assets}
    \item[] Question: Are new assets introduced in the paper well documented and is the documentation provided alongside the assets?
    \item[] Answer: \answerNA{} %
    \item[] Justification: NA

\item {\bf Crowdsourcing and research with human subjects}
    \item[] Question: For crowdsourcing experiments and research with human subjects, does the paper include the full text of instructions given to participants and screenshots, if applicable, as well as details about compensation (if any)? 
    \item[] Answer: \answerNA{} %
    \item[] Justification: NA

\item {\bf Institutional review board (IRB) approvals or equivalent for research with human subjects}
    \item[] Question: Does the paper describe potential risks incurred by study participants, whether such risks were disclosed to the subjects, and whether Institutional Review Board (IRB) approvals (or an equivalent approval/review based on the requirements of your country or institution) were obtained?
    \item[] Answer: \answerNA{} %
    \item[] Justification: NA

\item {\bf Declaration of LLM usage}
    \item[] Question: Does the paper describe the usage of LLMs if it is an important, original, or non-standard component of the core methods in this research? Note that if the LLM is used only for writing, editing, or formatting purposes and does not impact the core methodology, scientific rigorousness, or originality of the research, declaration is not required.
    \item[] Answer: \answerYes{} %
    \item[] Justification: MLLMs are core to the method, powering the agent, critic, and video tools being used, as discussed extensively throughout the paper.

\end{enumerate}

\bibliography{11_references.bib}
\bibliographystyle{plainnat}

\end{document}


\maketitle

\section{Module API}

The full module API can be seen in \ref{lst:module_api}. The modules included vary by the strategy being sampled.

\lstset{language=Python,
        basicstyle=\ttfamily\footnotesize,
        keywordstyle=\color{blue},
        commentstyle=\color{ForestGreen},
        stringstyle=\color{red},
        frame=single,
        breaklines=true,
        captionpos=b,
        numbersep=5pt}
\begin{lstlisting}[caption=Module API.,label=lst:module_api]
@dataclasses.dataclass
class VideoSegment:
  """Class containing a segment of a video, defined by start and end times as MM:SS strings."""
  start: str
  end: str

def think(thought: str) -> None:
  """Tool to perform intermediate reasoning that doesn't directly look at the video.

  Args:
    thought (str): the thought to print.

  Returns:
    None
  """

def get_segment(start: str, end: str) -> VideoSegment:
  """Clips the segment of the input video between the times indicated by `start` and `end`.

  Args:
    start (str): the start time for the segment as an MM:SS string.
    end (str): the end time for the segment as an MM:SS string.

  Returns:
    A VideoSegment made from the input video from `start` to `end`.
  """ 

def find_when(query: str, video_segment: VideoSegment | None) -> str:
  """Tool to determine timing of a query such as an event or action.
  Tries to find time ranges in the video that may correspond to the query. Works best for simple queries.
  The reasoning may be faulty, so you must carefully consider each time range and determine if it is relevant to best answering the question.

  Args:
    query (str): the event or action to localize.
    video_segment (VideoSegment): if specified, only looks at the given segment of the input video.
  Returns:
    A string of time ranges identified as potentially relevant and the justification for each time range.
    """

def asr_understanding(question: str, answer_options: list[str] | None):
  """Tool to understand the ASR transcript of the video. Considers the ASR transcript of the full video and tries to answer the question.
  If the question cannot be answered from the transcript, returns any potentially relevant information or timestamps based on the transcript.
  Args:
    question (str): the question to answer.
    answer_options (list[str]): Optional. If the set of possible answers is known, they should be specified here as a list.
  Returns:
    A string containing the predicted answer to the question.
  """

def retrieval_qa(question: str, answer_options: list[str] | None, video_segment: VideoSegment | None) -> str:
  """Tool to answer a question from frames selected independently from each minute of the video.
  Looks at each minute of the video and tries to identify relevant frames.
  Then tries to answer the question using the retrieved frames.

  Args:
    question (str): the question to answer.
    answer_options (list[str]): Optional. If the set of possible answers is known, they should be specified here as a list.
    video_segment (VideoSegment): if specified, only looks at the given segment of the input video.

  Returns:
    A string containing the predicted answer to the question.
  """
  
def finish(final_answer: str) -> str:
  """Print final answer and exit.
  After explaining your reasoning, output the final answer in the format "Final Answer: (X)" where X is the correct digit choice. Never say "unknown" or "unsure", or "None", instead provide your most likely guess.

  Args:
    final_answer (str): reasoning followed by final answer.

  Returns:
    The parsed final answer.
  """
\end{lstlisting}

\section{Module Implementation Details}

\texttt{VideoSegment}. This is a convenience class for identifying segments of the input video by their start and end time.

\texttt{get\_segment}. This module selects the appropriate segment from the input video given start and end times, returning a \texttt{VideoSegment} corresponding to that range.

\texttt{find\_when}. This corresponds to coarse temporal localization with a sliding window. Using a window of $100$ frames, the model is instructed to identify any time ranges that could be relevant for the given query, ultimately providing all of these ranges along with justifications for each of them as a string.

\texttt{asr\_understanding}. This module uses the ASR transcript of all the speech in the video with timestamps aligned for the words said at each second. If the query is a question that can be answered from the transcript, it tries to provide an answer. If not, it aims to provide any relevant information along with timestamps from the ASR given.

\texttt{retrieval\_qa}. This module tries to use the base model to retrieve frames that appear visually relevant to a query, then tries to answer the query from them. It uses a sliding window of $64$ frames, returning the indices of frames with high visual relevance from each window. Those frames are then retrieved and provided back to the base model, along with $56$ uniform frames from the rest of the video for additional context, to produce a response to the query.

\texttt{finish}. This module indicates the agent has reached a final answer and ends the agentic inference procedure, returning the final answer.

\lstset{
        basicstyle=\ttfamily\footnotesize,
        keywordstyle=\ttfamily,
        commentstyle=\ttfamily,
        stringstyle=\ttfamily,
        showstringspaces=false,
        frame=single,
        breaklines=true,
        captionpos=b,
        numbersep=5pt,
        keywords={}
        }
\begin{lstlisting}[caption=The preamble used for the reasoning agent.,label=lst:agentprompt]
You are a video understanding agent.  Your task is to answer the question provided using the available tools.  These tools are already imported and accessible to you. You will receive the question and answer choices.  You should use the tools iteratively, deciding which tool to use next based on the results of the previous tool call. Each turn, you should write a call to one tool, enclosed in triple backticks. Continue this process until you can confidently answer the question, selecting one of the provided answer choices.  Do not import any tools; they are already available.  Your final response should be one of the provided answer choices.

# Step by Step Instructions

1. **Analyze the question:** Carefully read the provided `question` to understand the specific information required.  Identify key entities, actions, and relationships mentioned in the question.

2. **Select an initial tool:** Based on your understanding of the question from step 1, choose the most appropriate tool from the available set of video interaction functions. Consider which tool will provide the most relevant information to answer the question efficiently. 

3. **Execute the selected tool:** Call the chosen tool, providing any necessary parameters derived from the question analysis in step 1.

4. **Analyze the tool's output:** Examine the results returned by the tool.  This output might be textual descriptions, numerical data, or other relevant information about the video.

5. **Evaluate progress:** Based on the tool's output from step 4, assess whether you have enough information to confidently answer the question. If yes, proceed to step 6. If not, go back to step 2 and select a different tool or use the same tool with different parameters based on the insights gained from the previous tool call.  Consider how the previous tool's output informs your choice of the next tool.

6. **Formulate the final answer:** Using the information gathered from all tool calls, select one of the provided answer choices that best answers the `question`.  Clearly state your chosen answer choice.

Tools:
\end{lstlisting}

\section{Strategy Sampling Details}

For each dataset, $3$ strategies are selected, each corresponding to a different subset of modules. These correspond to the strategies that can feasibly produce an answer for the given task, as well as the information allowed for the dataset (not including the base `think' and `finish' tools). For LVBench (and EgoSchema), the strategies are \texttt{get\_segment} and \texttt{retrieval\_qa}; \texttt{retrieval\_qa} directly; and \texttt{get\_segment}, \texttt{retrieval\_qa}, and \texttt{find\_when} (that is, all modules in use). For Neptune, as audio information is permitted, the same strategies are used with the addition of the \texttt{asr\_understanding} module for the multi-module strategies. For the RTL task, the strategies are \texttt{get\_segment} and \texttt{find\_when}; \texttt{find\_when} only; and \texttt{get\_segment}, \texttt{retrieval\_qa}, and \texttt{find\_when} (again, all modules in use).

\section{Critic Prompts}

The critic is given the following preamble.

\lstset{
        basicstyle=\ttfamily\footnotesize,
        keywordstyle=\ttfamily,
        commentstyle=\ttfamily,
        stringstyle=\ttfamily,
        showstringspaces=false,
        frame=single,
        breaklines=true,
        captionpos=b,
        numbersep=5pt,
        keywords={}
        }
\begin{lstlisting}[caption=The preamble used for the critic.,label=lst:criticpreamble]
You are an expert in assessing and critiquing reasoning about videos. Given a question and possible strategies from video reasoning agents for solving it, assess which strategy seems the most likely to result in a correct answer based on the provided examples.
If multiple strategies produce the same final answer, you may pick multiple; you may not pick multiple answers with conflicting final answers.
Observe that the reliability of the tools depends on their input; you must consider the examples to determine when a given tool is likely to hallucinate or give an accurate output, as evidenced by the ultimate result.
Produce a critique per the examples followed by "Winning Strategies:" followed by a comma separated list of the letters corresponding to those strategies.
\end{lstlisting}

The critic is given $4$ in-context examples per prompt. These examples are written by the authors with the aim of being in-distribution for the datasets shown. 

\lstset{
        basicstyle=\ttfamily\footnotesize,
        keywordstyle=\ttfamily,
        commentstyle=\ttfamily,
        stringstyle=\ttfamily,
        showstringspaces=false,
        frame=single,
        breaklines=true,
        captionpos=b,
        numbersep=5pt,
        keywords={}
        }
\begin{lstlisting}[caption=LVBench critic examples.,label=lst:lvbenchcriticex]
  Input:
  You will be given a question about a video and four possible answer options. Question: At 12:10, what color suit is the host wearing?Possible answer choices:
  (1) Navy
  (2) Tan
  (3) Black
  (4) White
  Video length: 40:50

  Strategy A (retrieval_qa, get_segment):
  ```
  input_video = get_segment(start='00:00', end='40:50')
  ```


  ```
  segment = get_segment(start='12:10', end='12:20')
  answer = retrieval_qa(question='What color suit is the host wearing?', video_segment=segment)
  ```
  The host is wearing a navy suit.

  ```
  final_answer = finish(f"The retrieval_qa tool determined the host was wearing a navy suit at 12:10. Final Answer: (1)")
  ```
  The retrieval_qa tool determined the host was wearing a navy suit at 12:10. Final Answer: (1)

  Strategy B (direct retrieval_qa):
  Here's how to solve this:

  1. **Locate the timestamp:** Go to the 12:10 mark in the video.

  2. **Observe the host's attire:** At this point, the host (Jimmy Kimmel) is wearing a dark suit and a dark-colored tie.

  3. **Compare to options:**  The suit appears to be black.

  Therefore, the correct answer is (3) Black.
  Final Answer: (3)

  Strategy C (retrieval_qa, get_segment, find_when):
  ```
  input_video = get_segment(start='00:00', end='40:50')
  ```


  ```
  segment = get_segment(start='12:10', end='12:20')
  ```
  ['12:10', '12:20']

  ```
  answer = retrieval_qa(question='At 12:10, what color suit is the host wearing?', answer_options=['Navy', 'Tan', 'Black', 'White'], video_segment=segment)
  ```
  The image at 12:10 shows the host wearing a navy suit.

  Final Answer: (1)

  ```
  final_answer = finish(final_answer='The image at 12:10 shows the host wearing a navy suit. Final Answer: (1)')
  ```
  The image at 12:10 shows the host wearing a navy suit. Final Answer: (1)

  Critique:
  A: 1. B: 3. C: 1. Strategies A and C succeed while Strategy B fails because Strategies A and C successfully use the get_segment tool to isolate the relevant part of the video.
  Strategy B fails as the retrieval_qa tool alone cannot actually locate a given timestamp in a video, resulting in a hallucinated answer.

  Winning Strategies:
  A, C

  Input:
  You will be given a question about a video and four possible answer options. Question: What does the man in charcoal suit do in the next scene after the scene of a photo of six people end?Possible answer choices:
  (1) He shows his scars
  (2) He takes off his shirt
  (3) He shows his jewelry
  (4) He shows his physique
  Video length: 39:59

  Strategy A (retrieval_qa, get_segment):
  ```
  input_video = get_segment(start='00:00', end='39:59')
  ```


  ```
  think("I need to identify the scene where a photo of six people is shown.  Then I need to find the scene immediately following it to see what the man in the charcoal suit does.")
  ```
  I need to identify the scene where a photo of six people is shown.  Then I need to find the scene immediately following it to see what the man in the charcoal suit does.

  ```
  segment_with_photo = get_segment(start='00:00', end='01:00')
  answer = retrieval_qa(question="What does the man in charcoal suit do in the scene immediately after the scene with a photo of six people?", video_segment=segment_with_photo)
  ```
  After the scene with the photo of six people, the man in the charcoal suit walks down a hallway and is greeted by a group of people who clap and cheer for him.

  ```
  think("The provided text doesn't describe any of the options (1) He shows his scars, (2) He takes off his shirt, (3) He shows his jewelry, (4) He shows his physique.  The man is celebrated.  I will guess the closest option, which is none of them, but I will choose the least violent option.")
  ```
  The provided text doesn't describe any of the options (1) He shows his scars, (2) He takes off his shirt, (3) He shows his jewelry, (4) He shows his physique.  The man is celebrated.  I will guess the closest option, which is none of them, but I will choose the least violent option.

  ```
  final_answer = finish("Final Answer: (1)")
  ```
  Final Answer: (1)

  Strategy B (direct retrieval_qa):
  Here's how to solve this step-by-step:

  1. **Identify the relevant scene:** The question refers to the scene after a photo of six people.  Looking through the timestamps, this appears to be around 20:49.

  2. **Observe the next scene:** The next scene shows the man in the charcoal suit (Harry Styles) pointing to and showing jewelry on his arm.

  3. **Match to the answer choices:** This action directly corresponds to answer choice (3).

  Therefore, the correct answer is **(3) He shows his jewelry**.
  Final Answer: (3)

  Strategy C (retrieval_qa, get_segment, find_when):
  ```
  input_video = get_segment(start='00:00', end='39:59')
  ```


  ```
  think("First, I need to find when the scene with the photo of six people ends.  Then I can examine the next scene to see what the man in the charcoal suit does.")
  ```
  First, I need to find when the scene with the photo of six people ends.  Then I can examine the next scene to see what the man in the charcoal suit does.

  ```
  photo_end_time = find_when(query="scene with a photo of six people", video_segment=None)
  ```
  ["00:05", "00:24"]: The query asks to find the scene with a photo of six people.  From the provided frames, a photo of six people is visible on the wall in the background starting at 00:05 and remaining visible until 00:24.  The photo is partially obscured at times by other people and objects, but it is consistently present in the background during this timeframe.
  ["10:39", "10:43"]: The query asks for a scene with a photo of six people.  At 10:39 through 10:43 a large group photo of more than six people is shown.
  ["17:47", "17:53"]: The query asks for a scene with a photo of six people.  At 17:47-17:53, a photo of approximately six people is visible behind a curtain as Harry Styles walks out onto the stage.  The photo is partially obscured and the exact number of people is difficult to ascertain with certainty, but it appears to show at least five.
  ["18:47", "19:59"]: The query asks for the time range of a scene with a photo of six people.  A photo of what appears to be six people is visible on the wall behind the main set in frames starting at 18:47 and continuing through 19:59.  The photo is partially obscured at times by other elements in the scene, but it is consistently visible enough to be identified.
  ["20:03", "20:04"]: The query asks for the time range of a scene with a photo of six people.  At 20:03, a wide shot of the studio audience is shown, and a photo of six people is visible on a screen in the background. This photo remains visible until approximately 20:04.
  ["23:15", "23:25"]: The query asks for the timestamp of a scene with a photo of six people.  In the provided video frames, a photo of six people is visible on the wall in the background starting at 23:15 and remaining visible until 23:25.  The photo is partially obscured at times by other elements in the frame, but it is consistently present during this time range.
  ["35:17", "35:20"]: The provided video frames show a wide shot of a studio audience at the 35:17 mark.  The audience appears to consist of more than six people. While the exact number is difficult to determine from the provided frames, it is clearly a large group. Therefore, this time range satisfies the query.
  ["39:03", "39:18"]: The provided video frames show a live performance with an audience.  At the end of the performance, from 39:03 to 39:18, the camera focuses on the audience, and there are enough people visible to potentially include a group of five.  The resolution isn't high enough to definitively confirm the presence of a specific photo of six people, but the scene depicts a large group of people, making it a plausible match for the query.

  ```
  think("The most likely candidate for the 'scene with a photo of six people' is the longest continuous segment: 18:47-19:59. Let's use that as the end point for the previous scene.")
  ```
  The most likely candidate for the 'scene with a photo of six people' is the longest continuous segment: 18:47-19:59. Let's use that as the end point for the previous scene.

  ```
  next_scene_segment = get_segment(start='19:59', end='20:05')
  ```
  ['19:59', '20:05']

  ```
  answer = retrieval_qa(question="What does the man in the charcoal suit do in the scene from 19:59 to 20:05?", answer_options=['He shows his scars', 'He takes off his shirt', 'He shows his jewelry', 'He shows his physique'], video_segment=next_scene_segment)
  ```
  The provided video frames show a talk show segment featuring three men.  From 19:59 to 20:02, the man in the gray suit (Harry Styles) sits on a couch and gestures with his hands.  At 20:03, the camera angle changes to show the audience, and the three men are still on the couch.  There is no indication that the man in the gray suit shows his scars, takes off his shirt, shows a tattoo, or shows his muscles.  He is simply participating in a conversation.

  Final Answer: (4)

  ```
  final_answer = finish("Based on the analysis of the video segment from 19:59 to 20:05, the man in the charcoal suit (assuming this refers to one of the men in the talk show segment, as there's no charcoal suit explicitly mentioned) does not perform any of the actions listed in the options. However, since an answer must be chosen, and the retrieval_qa tool did not provide a definitive answer, I will select the option that is most likely to be present in a talk show setting, which is showing his physique.  This is a weak inference, but given the constraints, it's the best guess. Final Answer: (4)")
  ```
  Based on the analysis of the video segment from 19:59 to 20:05, the man in the charcoal suit (assuming this refers to one of the men in the talk show segment, as there's no charcoal suit explicitly mentioned) does not perform any of the actions listed in the options. However, since an answer must be chosen, and the retrieval_qa tool did not provide a definitive answer, I will select the option that is most likely to be present in a talk show setting, which is showing his physique.  This is a weak inference, but given the constraints, it's the best guess. Final Answer: (4)

  Critique:
  A:1. B: 3. C: 4. Strategy B succeeds while Strategies A fails because Strategy A incorrectly applies the get_segment tool to the first minute of the video and only considers that, resulting in it not being able to find a correct answer and ultimately guessing. Strategy C applies the find_when tool, then trims a segment and does not find the relevant information, then guessing as well.

  Winning Strategies:
  B

  Input:
  You will be given a question about a video and four possible answer options. Question: What is the main reason the vlogger goes out again, where she also buys makeup?Possible answer choices:
  (1) To go exercise
  (2) To eat chicken
  (3) To buy coffee mugs
  (4) To go to the park
  Video length: 30:25

  Strategy A (retrieval_qa, get_segment):
  ```
  input_video = get_segment(start='00:00', end='30:25')
  ```


  ```
  think("I need to find a segment of the video where the vlogger buys makeup.  I'll try to identify this segment using retrieval_qa, and then I can use that information to answer the question about why she went out.")
  ```
  I need to find a segment of the video where the vlogger buys makeup.  I'll try to identify this segment using retrieval_qa, and then I can use that information to answer the question about why she went out.

  ```
  answer = retrieval_qa(question="When does the vlogger buy makeup?", answer_options=None, video_segment=input_video)
  ```
  The vlogger buys makeup at 16:10-16:11 and 17:20-17:25.

  ```
  think("The vlogger buys makeup in two segments: 16:10-16:11 and 17:20-17:25.  I'll use retrieval_qa again, focusing on these segments, to determine the reason for her outing.")
  ```
  The vlogger buys makeup in two segments: 16:10-16:11 and 17:20-17:25.  I'll use retrieval_qa again, focusing on these segments, to determine the reason for her outing.

  ```
  segment1 = get_segment(start='16:10', end='16:11')
  segment2 = get_segment(start='17:20', end='17:25')
  reason = retrieval_qa(question="What is the main reason the vlogger goes out again, where she also buys makeup?", answer_options=['To go exercise', 'To eat chicken', 'To buy coffee mugs', 'To go to the park'], video_segment=segment1)
  if reason == "": #check if the first segment didn't provide enough information
      reason = retrieval_qa(question="What is the main reason the vlogger goes out again, where she also buys makeup?", answer_options=['To go exercise', 'To eat chicken', 'To buy coffee mugs', 'To go to the park'], video_segment=segment2)
  ```
  None

  ```
  final_answer = finish("Based on the analysis of the video segments where the vlogger buys makeup, I was unable to definitively determine the main reason for her outing from the provided options. However, given the context of beauty product purchases, it's less likely she's going to the gym, eating salmon, or going to the park. Therefore, I will eliminate those options.  Final Answer: (3)")
  ```
  Based on the analysis of the video segments where the vlogger buys makeup, I was unable to definitively determine the main reason for her outing from the provided options. However, given the context of beauty product purchases, it's less likely she's going to the gym, eating salmon, or going to the park. Therefore, I will eliminate those options.  Final Answer: (3)

  Strategy B (direct retrieval_qa):
  Here's how to solve this step-by-step:

  1. **Identify the relevant section:** The question asks about the vlogger's second outing, where she buys makeup.  This occurs after the gym visit.

  2. **Review the frames:** The frames show the vlogger going to the park after the gym.  She then goes shopping and buys makeup at Kicks.

  3. **Select the answer:** The main reason for her second outing is to go to the park.

  Therefore, the correct answer is (4) To go to the park.
  Final Answer: (4)

  Strategy C (retrieval_qa, get_segment, find_when):
  ```
  input_video = get_segment(start='00:00', end='30:25')
  ```


  ```
  think("I need to find when the vlogger goes out and buys makeup.  I'll use `find_when` to locate that event, then use `retrieval_qa` to determine the reason.")
  ```
  I need to find when the vlogger goes out and buys makeup.  I'll use `find_when` to locate that event, then use `retrieval_qa` to determine the reason.

  ```
  when_out_buying_makeup = find_when(query="vlogger goes out and buys makeup", video_segment=input_video)
  ```
  ["00:01", "01:59"]: The video shows a vlogger waking up in her room, doing some laundry, and then seemingly getting ready to go out.  While there's no explicit depiction of buying makeup, the context strongly suggests this is part of her preparation. The time range encompasses her actions from waking up to preparing laundry, which could be interpreted as part of her routine before going out to buy makeup.
  ["02:19", "02:25"]: The video shows the vlogger in her bathroom at 02:15, where she picks up a toothbrush and looks at her tablet.  From 02:19-02:25, she is shown selecting items from a bathroom cabinet. This is consistent with the query.
  ["05:24", "05:31"]: The vlogger is seen leaving her apartment at 04:02 and returning at 04:48.  At 05:24-05:31, she is seen putting away what appears to be a package of makeup on a shelf in her bathroom. This is supported by the context of the query and the visual of a package being placed on a shelf.
  ["08:51", "09:16"]: The video shows a vlogger leaving their home at 08:30, walking outside, and then entering a grocery store at 08:51.  The vlogger is seen selecting various grocery items, including celery, grapes, blueberries, carrots, broccoli, and lettuce, between 08:51 and 09:16.  Therefore, the time range reflects the vlogger's grocery shopping trip, which aligns with the query's description of buying products.
  ["14:00", "14:02"]: The query asks about the vlogger buying makeup.  The provided video frames show the vlogger in their home, interacting with items that could be interpreted as makeup (frames 14:00-14:02) and later organizing clothes in their closet (frames 14:39-15:25). There is no visual evidence of the vlogger going out to buy anything. Therefore, only the initial interaction with what appears to be makeup is considered relevant.
  ["16:09", "16:18"], ["17:17", "17:25"]: The video shows the vlogger going to a store, Kicks, and buying makeup.  The vlogger is seen in the store from 16:09 to 16:18. At 17:17-17:25, the vlogger unpacks the Kicks bag, revealing makeup.
  ["18:12", "18:17"]: The query asks for the time range when the vlogger goes out and buys makeup.  The provided video frames show the vlogger in their room, at their vanity. At 18:12, there are shopping bags from the store Kicks on the vanity. This strongly suggests that the vlogger has recently returned from a shopping trip where they purchased makeup from Kicks.  There is no footage of the actual shopping trip itself within this video chunk.
  ["20:25", "20:51"]: The video shows a person preparing coffee.  From 20:25 to 20:51, the person is shown unpacking two ceramic brown mugs. The contents of the mugs are visible and appear to be coffee.  Therefore, this time range is identified as the time when the vlogger is shown preparing coffee.
  ["29:20", "29:41"]: The video shows the vlogger in her bathroom at 28:00-28:18 applying makeup and then brushing her teeth.  At 29:20-29:41, she is shown painting her nails.  The presence of various makeup in both locations supports the query.

  ```
  reason = retrieval_qa(question="What is the main reason the vlogger goes out again, where she also buys makeup?", answer_options=['To go exercise', 'To eat chicken', 'To buy coffee mugs', 'To go to the park'], video_segment=get_segment(start='16:09', end='17:25'))
  ```
  The vlogger goes to a store and buys coffee mugs. Then she goes to another store and buys makeup. Therefore, the main reason she goes out again is to buy coffee mugs.

  Final Answer: (3)

  ```
  final_answer = finish("The provided text from `retrieval_qa` indicates the vlogger buys makeup at a store called Kicks.  However, it also mentions buying coffee mugs. The question asks for the *main* reason for going out again. While makeup is mentioned, the provided text strongly suggests the primary reason for the second outing is to buy coffee mugs. Therefore, the answer is (3). Final Answer: (3)")
  ```
  The provided text from `retrieval_qa` indicates the vlogger buys makeup at a store called Kicks.  However, it also mentions buying coffee mugs. The question asks for the *main* reason for going out again. While makeup is mentioned, the provided text strongly suggests the primary reason for the second outing is to buy coffee mugs. Therefore, the answer is (3). Final Answer: (3)

  Critique:
  A: 3. B: 4. C: 3.
  Strategy B succeeds while Strategies A fails because Strategy A looks at particular time ranges of the video and concludes it is unable to find a correct answer, thus guessing.
  While the approach is generally sound, Strategy A gives up upon not finding an answer initially. Strategy C looks at a narrow range and follows an implication to guess an answer. Thus Strategy B is preferred.

  Winning Strategy:
  B
\end{lstlisting}

\lstset{
        basicstyle=\ttfamily\footnotesize,
        keywordstyle=\ttfamily,
        commentstyle=\ttfamily,
        stringstyle=\ttfamily,
        showstringspaces=false,
        frame=single,
        breaklines=true,
        captionpos=b,
        numbersep=5pt,
        keywords={}
        }
\begin{lstlisting}[caption=Neptune critic examples.,label=lst:neptunecriticex]
Input:
You will be given a question about a video and five possible answer options. Question:  What are the main points regarding proper form and breathing during the barbell squat? Possible answer choices:
(1)  Keep your back straight, core engaged, and chest up throughout the movement. Inhale as you descend and exhale forcefully as you drive up.
(2)  Maintain a slight arch in your back, push your hips back first, and keep your knees aligned with your toes. Breathe in during the eccentric phase (lowering) and breathe out during the concentric phase (rising).
(3)  Focus on pushing through your heels, keep your head down, and hold your breath at the bottom of the squat for stability. Exhale as you return to the starting position.
(4)  Prioritize depth over weight, allowing your hips to descend below your knees. Inhale deeply at the top, hold your breath throughout the squat, and exhale at the top.
(5)  Use a wide stance, keep your toes pointed outwards, and focus on squeezing your glutes at the top of the movement. Breathe naturally throughout the exercise.
Video length: 01:35


Strategy A (get_segment, retrieval_qa, asr_understanding):
```
input_video = get_segment(start='00:00', end='01:35')
```


```
think("I need to understand the main points about form and breathing during barbell squats.  I'll use the ASR understanding tool to analyze the video transcript. If that doesn't work, I'll try `retrieval_qa` on the video frames.")
```
I need to understand the main points about form and breathing during barbell squats.  I'll use the ASR understanding tool to analyze the video transcript. If that doesn't work, I'll try `retrieval_qa` on the video frames.

```
asr_understanding(question='What are the main points regarding proper form and breathing during the barbell squat?', answer_options=['(1)  Keep your back straight, core engaged, and chest up throughout the movement. Inhale as you descend and exhale forcefully as you drive up.', '(2)  Maintain a slight arch in your back, push your hips back first, and keep your knees aligned with your toes. Breathe in during the eccentric phase (lowering) and breathe out during the concentric phase (rising).', '(3)  Focus on pushing through your heels, keep your head down, and hold your breath at the bottom of the squat for stability. Exhale as you return to the starting position.', '(4)  Prioritize depth over weight, allowing your hips to descend below your knees. Inhale deeply at the top, hold your breath throughout the squat, and exhale at the top.', '(5)  Use a wide stance, keep your toes pointed outwards, and focus on squeezing your glutes at the top of the movement. Breathe naturally throughout the exercise.'])
```
The video emphasizes maintaining a slight arch in the back (00:23-00:27), pushing the hips back first (00:28-00:31), and ensuring knees track with the toes (00:32-00:35). For breathing, it advises inhaling during the descent and exhaling during the ascent (00:55-01:00).

```
think("The ASR transcript supports answer choice (2). The video highlights maintaining a slight arch in your back, initiating the movement with the hips, aligning knees with toes, and coordinating breathing with the phases of the squat.")
```
The ASR transcript supports answer choice (2). The video highlights maintaining a slight arch in your back, initiating the movement with the hips, aligning knees with toes, and coordinating breathing with the phases of the squat.

```
final_answer = finish("Final Answer: (2)")
```
Final Answer: (2)

Strategy B (direct retrieval_qa):
The video primarily shows a demonstration of the barbell squat and highlights key points on form.  It doesn't explicitly detail the breathing technique. However, general weightlifting principles suggest inhaling during the lowering phase and exhaling during the lifting phase. Additionally, proper form includes keeping the back straight, core engaged, and chest up.


Final Answer: (1)
Strategy C (get_segment, retrieval_qa find_when, asr_understanding):
```
input_video = get_segment(start='00:00', end='01:35')
```


```
think("I need to find the main points regarding form and breathing in the barbell squat.  I'll use `asr_understanding` to see if the transcript mentions these details.")
```
I need to find the main points regarding form and breathing in the barbell squat.  I'll use `asr_understanding` to see if the transcript mentions these details.

```
asr_understanding_result = asr_understanding(question="What are the main points regarding proper form and breathing during the barbell squat?", answer_options=[
    "Keep your back straight, core engaged, and chest up throughout the movement. Inhale as you descend and exhale forcefully as you drive up.",
    "Maintain a slight arch in your back, push your hips back first, and keep your knees aligned with your toes. Breathe in during the eccentric phase (lowering) and breathe out during the concentric phase (rising).",
    "Focus on pushing through your heels, keep your head down, and hold your breath at the bottom of the squat for stability. Exhale as you return to the starting position.",
    "Prioritize depth over weight, allowing your hips to descend below your knees. Inhale deeply at the top, hold your breath throughout the squat, and exhale at the top.",
    "Use a wide stance, keep your toes pointed outwards, and focus on squeezing your glutes at the top of the movement. Breathe naturally throughout the exercise."
])
```
The video focuses on maintaining a slight arch in your back (0:23-0:27), pushing your hips back first (0:28-0:31), and keeping your knees aligned with your toes (0:32-0:35). For breathing, it advises to inhale during the descent (eccentric phase) and exhale during the ascent (concentric phase) (0:55-1:00).

```
think("The transcript suggests answer (2), highlighting back arch, hip movement, knee alignment, and breathing coordinated with the squat phases.  Let's use `retrieval_qa` to cross-check this.")
```
The transcript suggests answer (2), highlighting back arch, hip movement, knee alignment, and breathing coordinated with the squat phases.  Let's use `retrieval_qa` to cross-check this.

```
retrieval_qa_result = retrieval_qa(question="What are the main points regarding proper form and breathing during the barbell squat?", answer_options=[
    "Keep your back straight, core engaged, and chest up throughout the movement. Inhale as you descend and exhale forcefully as you drive up.",
    "Maintain a slight arch in your back, push your hips back first, and keep your knees aligned with your toes. Breathe in during the eccentric phase (lowering) and breathe out during the concentric phase (rising).",
    "Focus on pushing through your heels, keep your head down, and hold your breath at the bottom of the squat for stability. Exhale as you return to the starting position.",
    "Prioritize depth over weight, allowing your hips to descend below your knees. Inhale deeply at the top, hold your breath throughout the squat, and exhale at the top.",
    "Use a wide stance, keep your toes pointed outwards, and focus on squeezing your glutes at the top of the movement. Breathe naturally throughout the exercise."
], video_segment=input_video)
```
The video demonstrates proper form for barbell squats, emphasizing keeping the back straight, core engaged, and chest up.  While breathing is not explicitly addressed, general knowledge suggests inhaling on the descent and exhaling forcefully on the ascent.

Final Answer: (1)

```
final_answer = finish("The ASR transcript suggests (2), focusing on back arch and hip movement. However, `retrieval_qa` prioritizes a straight back and doesn't fully confirm the breathing technique. Considering general weightlifting knowledge and the potential for ASR misinterpretation, (1) is the more reliable answer. Final Answer: (1)")
```
The ASR transcript suggests (2), focusing on back arch and hip movement. However, `retrieval_qa` prioritizes a straight back and doesn't fully confirm the breathing technique. Considering general weightlifting knowledge and the potential for ASR misinterpretation, (1) is the more reliable answer. Final Answer: (1)

Critique:
A: 2. B: 1. C: 1. Strategies B and C consider the visual content, but deny the information can be found in the video and rely on general knowledge.

Winning Strategies:
A

Input:
You will be given a question about a video and five possible answer options. Question: To prepare this dish, what are the main steps involved?Possible answer choices:
(1) The main steps include chopping vegetables, boiling them in water, adding spices and herbs, pouring in coconut milk, simmering the mixture, and finally garnishing with cilantro.
(2) The main steps include chopping vegetables, sauteing them in a skillet, adding spices and dry fruits, pouring in beef broth, baking the mixture in the oven, and finally garnishing with green and red bell peppers.
(3) The main steps include chopping vegetables, sauteing them in a wok, adding milk and herbs, pouring in vegetable broth, simmering the mixture, and finally garnishing with parsley.
(4) The main steps include chopping vegetables, sauteing them in a pan, adding lentils, pouring in water, adding spices and cilantro, simmering the lentils, and finally garnishing with chopped tomatoes.
(5) The main steps include chopping fruits, sauteing them in a pan, adding spices and herbs, pouring in chicken broth, simmering the mixture, and finally garnishing with cilantro.
Video length: 09:43

Strategy A (get_segment, retrieval_qa, asr_understanding):
```
input_video = get_segment(start='00:00', end='09:43')
```


```
answer = asr_understanding(question='To prepare this dish, what are the main steps involved?', answer_options=['The main steps include chopping vegetables, boiling them in water, adding spices and herbs, pouring in coconut milk, simmering the mixture, and finally garnishing with cilantro.', 'The main steps include chopping vegetables, sauteing them in a skillet, adding spices and dry fruits, pouring in beef broth, baking the mixture in the oven, and finally garnishing with green and red bell peppers.', 'The main steps include chopping vegetables, sauteing them in a wok, adding milk and herbs, pouring in vegetable broth, simmering the mixture, and finally garnishing with parsley.', 'The main steps include chopping vegetables, sauteing them in a pan, adding lentils, pouring in water, adding spices and cilantro, simmering the lentils, and finally garnishing with chopped tomatoes.', 'The main steps include chopping fruits, sauteing them in a pan, adding spices and herbs, pouring in chicken broth, simmering the mixture, and finally garnishing with cilantro.'])
```
The main steps are chopping onions and carrots (00:32-01:15), sauteing them in olive oil with garlic and ginger (01:23-02:58), adding lentils and water (03:12-04:05), adding turmeric and cumin (04:30-05:20), simmering until the lentils are tender (05:35-07:10), and finally garnishing with chopped cilantro and tomatoes (07:22-08:01).

```
think("None of the provided options exactly match the steps in the ASR transcript. Option (1) is closest since it mentions chopping vegetables, adding spices, and simmering. However, it omits key ingredients and steps like sauteing.")
```
None of the provided options exactly match the steps in the ASR transcript. Option (1) is closest since it mentions chopping vegetables, adding spices, and simmering. However, it omits key ingredients and steps like sauteing.

```
final_answer = finish("Final Answer: (1)")
```
Final Answer: (1)

Strategy B (direct retrieval_qa):
The video demonstrates a recipe for lentil soup.  The chef chops vegetables like onions, carrots, and garlic. They then saute these vegetables in a pan with oil. Lentils and water are added, followed by spices like cumin and turmeric. The mixture simmers until the lentils are cooked through. Finally, it's garnished with chopped tomatoes and cilantro.

The closest answer is (4).

Final Answer: (4)
Strategy C (get_segment, retrieval_qa find_when, asr_understanding):
```
input_video = get_segment(start='00:00', end='09:43')
```


```
answer = asr_understanding(question='To prepare this dish, what are the main steps involved?', answer_options=['The main steps include chopping vegetables, boiling them in water, adding spices and herbs, pouring in coconut milk, simmering the mixture, and finally garnishing with cilantro.', 'The main steps include chopping vegetables, sauteing them in a skillet, adding spices and dry fruits, pouring in beef broth, baking the mixture in the oven, and finally garnishing with green and red bell peppers.', 'The main steps include chopping vegetables, sauteing them in a wok, adding milk and herbs, pouring in vegetable broth, simmering the mixture, and finally garnishing with parsley.', 'The main steps include chopping vegetables, sauteing them in a pan, adding lentils, pouring in water, adding spices and cilantro, simmering the lentils, and finally garnishing with chopped tomatoes.', 'The main steps include chopping fruits, sauteing them in a pan, adding spices and herbs, pouring in chicken broth, simmering the mixture, and finally garnishing with cilantro.'])
```
The main steps include chopping onions and carrots (0:35-1:10), sauteing them in olive oil (1:15-2:05), adding spices like cumin and turmeric (2:10-2:45), then adding lentils and water (2:50-3:30). The mixture simmers for some time (3:45-5:00) and is finally garnished with cilantro and chopped tomato (5:10-5:45).

```
think("None of the options perfectly align with the ASR transcript. Option (1) is somewhat similar as it mentions chopping vegetables, spices, and simmering. However, it misses crucial details like sauteing and the specific ingredients mentioned.")
```
None of the options perfectly align with the ASR transcript. Option (1) is somewhat similar as it mentions chopping vegetables, spices, and simmering. However, it misses crucial details like sauteing and the specific ingredients mentioned.

```
final_answer = finish("Final Answer: (1)")
```
Final Answer: (1)

Critique:
A: 1. B: 4. C: 1. A and C consider the ASR, but give up after finding it doesn't directly result in an answer.

Winning Strategies:
B

Input:
You will be given a question about a video and five possible answer options. Question: In what order does the video discuss the following topics: introduction, stock taillight operation, Morimoto XB LED taillight replacement, taillight features, and installation and warranty? Possible answer choices:
(1) Introduction, stock taillight operation, taillight features, Morimoto XB LED taillight replacement, installation and warranty.
(2) Morimoto XB LED taillight replacement, introduction, taillight features, stock taillight operation, installation and warranty.
(3) Introduction, Stock taillight operation, Morimoto XB LED taillight replacement, taillight features, Installation and warranty.
(4) Introduction, Morimoto XB LED taillight replacement, installation and warranty, taillight features, stock taillight operation.
(5) Stock taillight operation, introduction, Morimoto XB LED taillight replacement, taillight features, installation and warranty.
Video length: 01:15

Strategy A (get_segment, retrieval_qa, asr_understanding):
```
input_video = get_segment(start='00:00', end='01:15')
```


```
answer = asr_understanding(question='In what order does the video discuss the following topics: introduction, stock taillight operation, Morimoto XB LED taillight replacement, taillight features, and installation and warranty?', answer_options=['Introduction, stock taillight operation, taillight features, Morimoto XB LED taillight replacement, installation and warranty.', 'Morimoto XB LED taillight replacement, introduction, taillight features, stock taillight operation, installation and warranty.', 'Introduction, Stock taillight operation, Morimoto XB LED taillight replacement, taillight features, Installation and warranty.', 'Introduction, Morimoto XB LED taillight replacement, installation and warranty, taillight features, stock taillight operation.', 'Stock taillight operation, introduction, Morimoto XB LED taillight replacement, taillight features, installation and warranty.'])
```
The video first provides an introduction (00:02-00:10).  Next, it discusses the stock taillight operation (00:11-00:25).  Then, it details the Morimoto XB LED taillight replacement (00:26-00:38).  Following this, the video describes the taillight features (00:39-00:55). Finally, it covers installation and warranty (00:56-01:08).  Therefore, the correct order is (3).

```
final_answer = finish("Final Answer: (3)")
```
Final Answer: (3)

Strategy B (direct retrieval_qa):
The video begins with an introduction of Morimoto and then shows the stock taillight operation. After that, it shows the Morimoto XB LED taillight replacement and then discusses its features. Finally, it shows the installation process and discusses its warranty. Therefore, the correct order is introduction, stock taillight operation, Morimoto XB LED taillight replacement, taillight features, and installation and warranty.


Final Answer: (1)
Strategy C (get_segment, retrieval_qa find_when, asr_understanding):
```
input_video = get_segment(start='00:00', end='01:15')
```


```
answer = asr_understanding(question='In what order does the video discuss the following topics: introduction, stock taillight operation, Morimoto XB LED taillight replacement, taillight features, and installation and warranty?', answer_options=['Introduction, stock taillight operation, taillight features, Morimoto XB LED taillight replacement, installation and warranty.', 'Morimoto XB LED taillight replacement, introduction, taillight features, stock taillight operation, installation and warranty.', 'Introduction, Stock taillight operation, Morimoto XB LED taillight replacement, taillight features, Installation and warranty.', 'Introduction, Morimoto XB LED taillight replacement, installation and warranty, taillight features, stock taillight operation.', 'Stock taillight operation, introduction, Morimoto XB LED taillight replacement, taillight features, installation and warranty.'])
```
The video first provides an introduction (00:02-00:10).  Next, it discusses the stock taillight operation (00:11-00:25).  Then, it details the Morimoto XB LED taillight replacement (00:26-00:38).  Following this, the video describes the taillight features (00:39-00:55). Finally, it covers installation and warranty (00:56-01:08).  Therefore, the correct order is (3).

```
final_answer = finish("Final Answer: (3)")
```
Final Answer: (3)

Critique:
A: 3. B: 1. C: 3. The question asks about what is discussed in the video, implying speech, but B is not able to consider the ASR and fails.

Winning Strategies:
A, C

Input:
You will be given a question about a video and five possible answer options. Question: Why does Jamie decide to stay home instead of going to the library at 11:00 a.m. on a Saturday? Possible answer choices:
(1) Jamie gets a call from a friend and decides to meet them for lunch instead.
(2) Jamie realizes the library is closed on Saturdays.
(3) Jamie's car breaks down, preventing them from driving to the library.
(4) Jamie starts a new book at home and is too engrossed to leave.
(5) Jamie remembers they have a prior engagement, a family gathering, preventing them from going to the library.
Video length: 01:59

Strategy A (get_segment, retrieval_qa, asr_understanding):
```
input_video = get_segment(start='00:00', end='01:59')
```


```
answer = asr_understanding(question="Why does Jamie decide to stay home instead of going to the library at 11:00 a.m. on a Saturday?", answer_options=['Jamie gets a call from a friend and decides to meet them for lunch instead.', 'Jamie realizes the library is closed on Saturdays.', 'Jamie\'s car breaks down, preventing them from driving to the library.', 'Jamie starts a new book at home and is too engrossed to leave.', 'Jamie remembers they have a prior engagement, a family gathering, preventing them from going to the library.'])
```
Jamie decides to stay home instead of going to the library because they remember they have a prior family gathering [01:25:00-01:32:00].

```
final_answer = finish("Based on the ASR transcript, Jamie decides to stay home instead of going to the library because they remember they have a prior family gathering. Final Answer: (5)")
```
Based on the ASR transcript, Jamie decides to stay home instead of going to the library because they remember they have a prior family gathering. Final Answer: (5)

Strategy B (direct retrieval_qa):
The video shows Jamie checking the time on their watch, which reads 11:00 AM. They then pack a bag with books and prepare to leave the house, even putting on their shoes. However, they receive a text message and after reading it, unpack their bag and take off their shoes. They then proceed to sit on the couch and relax. The video doesn't explicitly say why the plans changed, but the text message is likely the reason.  The video ends with a short scene of a group of people, suggesting a possible social gathering.


Final Answer: (1)

Strategy C (get_segment, retrieval_qa find_when, asr_understanding):
```
input_video = get_segment(start='00:00', end='01:59')
```


```
found_times = find_when(query="Jamie's plans are interrupted", video_segment=input_video)
```
["00:35", "00:45"]: The query asks for the time range when Jamie's plans are interrupted.  Based on the video and transcript, Jamie's plans to go to the library are interrupted when they trip over their shoelaces after putting on their shoes. This occurs between 00:35 and 00:42. The video shows them putting on their shoes, tripping, and then sitting down.

```
segment_of_interest = get_segment(start='00:35', end='00:45')
```
['00:35', '00:45']

```
answer = retrieval_qa(question="Why does Jamie decide to stay home instead of going to the library at 11:00 a.m. on a Saturday?", answer_options=['Jamie gets a call from a friend and decides to meet them for lunch instead.', 'Jamie realizes the library is closed on Saturdays.', 'Jamie\'s car breaks down, preventing them from driving to the library.', 'Jamie starts a new book at home and is too engrossed to leave.', 'Jamie remembers they have a prior engagement, a family gathering, preventing them from going to the library.'], video_segment=segment_of_interest)
```
The video shows Jamie putting on their shoes and then tripping over their shoelaces.  There is no indication they got a call, realized the library was closed, had car trouble, or started a new book. The video also does not show a family gathering.  Therefore, none of the options 1-5 are supported by the video.

However, if we must choose the most likely option, option (3) is the most plausible. Jamie tripping could be interpreted as an indication of car trouble, albeit a very indirect one.

Final Answer: (3)

```
final_answer = finish(final_answer="The video shows Jamie putting on their shoes and then tripping over their shoelaces.  There is no indication they got a call, realized the library was closed, had car trouble, or started a new book. The video also does not show a family gathering.  Therefore, none of the options 1-5 are supported by the video. However, if we must choose the most likely option, option (3) is the most plausible. Jamie tripping could be interpreted as an indication of car trouble, albeit a very indirect one. Final Answer: (3)")
```
The video shows Jamie putting on their shoes and then tripping over their shoelaces.  There is no indication they got a call, realized the library was closed, had car trouble, or started a new book. The video also does not show a family gathering.  Therefore, none of the options 1-5 are supported by the video. However, if we must choose the most likely option, option (3) is the most plausible. Jamie tripping could be interpreted as an indication of car trouble, albeit a very indirect one. Final Answer: (3)

Critique:
A: 5. B: 1. C: 3. B makes a guess after not finding an explicit answer in the visuals. C selects an overly narrow range to consider, as the reason for the plans being interrupted may not be exactly when they are interrupted.

Winning Strategies:
A
\end{lstlisting}

\lstset{
        basicstyle=\ttfamily\footnotesize,
        keywordstyle=\ttfamily,
        commentstyle=\ttfamily,
        stringstyle=\ttfamily,
        showstringspaces=false,
        frame=single,
        breaklines=true,
        captionpos=b,
        numbersep=5pt,
        keywords={}
        }
\begin{lstlisting}[caption=ActivityNet-RTL critic examples.,label=lst:rtlcriticex]
Input:
Question: When is the trampoline first used in the video?
Video length: 04:23

Strategy A (get_segment, find_when):
```
input_video = get_segment(start='00:00', end='04:23')
```


```
time_ranges = find_when(query='the trampoline gets its first use', video_segment=input_video)
```
["00:45", "01:15"]: The video shows a person assembling a trampoline. The trampoline is shown fully assembled at 00:45 and the assembly process is shown up to 01:15.
["02:30", "02:55"]: The video shows a person jumping on the trampoline. This is the first time the trampoline is shown being used, from 02:30 to 02:55.

```
think("The first time the trampoline is used is arguably at 00:45, when it is assembled.  However, the question asks when it gets its *first use*, which implies actual use by a person. Therefore, the relevant time range is 02:30-02:55.")
```
The first time the trampoline is used is arguably at 00:45, when it is assembled.  However, the question asks when it gets its *first use*, which implies actual use by a person. Therefore, the relevant time range is 02:30-02:55.

```
final_answer = finish("The question asks when the trampoline gets its first use, implying actual use by a person. While the trampoline is assembled at 00:45-01:15, its first actual use begins at 02:30 when a person jumps on it. Therefore, the relevant time range is 02:30-02:55. Final Answer: [02:30, 02:55]")
```
The question asks when the trampoline gets its first use, implying actual use by a person. While the trampoline is assembled at 00:45-01:15, its first actual use begins at 02:30 when a person jumps on it. Therefore, the relevant time range is 02:30-02:55. Final Answer: [02:30, 02:55]

Strategy B (get_segment, find_when, retrieval_qa):
```
input_video = get_segment(start='00:00', end='04:23')
```


```
first_use = find_when(query='first use of the trampoline', video_segment=input_video)
```
["00:45", "01:50"]: The video shows a person assembling a trampoline. The first appearance of the trampoline occurs at 00:45, when it's partially assembled.  The trampoline is shown in several stages of assembly until 01:50.
["03:10", "03:40"]: The video shows the finished trampoline being showcased in a backyard. The trampoline is fully assembled and ready for use at 03:10, and is shown from various angles until 03:40.

```
think("The first time range [00:45, 01:50] seems to describe the trampoline being assembled, but the second range [03:10, 03:40] shows the trampoline in its final assembled state, ready for use.  Therefore, the answer is the second range.")
```
The first time range [00:45, 01:50] seems to describe the trampoline being assembled, but the second range [03:10, 03:40] shows the trampoline in its final assembled state, ready for use.  Therefore, the answer is the second range.

```
final_answer = finish("The first time range [00:45, 01:50] seems to describe the trampoline being assembled, but the second range [03:10, 03:40] shows the trampoline in its final assembled state, ready for use.  Therefore, the answer is the second range. Final Answer: [03:10, 03:40]")
```
The first time range [00:45, 01:50] seems to describe the trampoline being assembled, but the second range [03:10, 03:40] shows the trampoline in its final assembled state, ready for use.  Therefore, the answer is the second range. Final Answer: [03:10, 03:40]

Strategy C (find_when):
```
input_video = get_segment(start='00:00', end='03:53')
```


```
query = "When is the trampoline first used in the video?"
result = find_when(query, None)
```
["00:45", "01:15"]: The video shows a person assembling a trampoline. The trampoline is shown fully assembled at 00:45 and the assembly process is shown up to 01:15.
["02:30", "02:55"]: The video shows a person jumping on the trampoline. This is the first time the trampoline is shown being used, from 02:30 to 02:55.

```
think("The first result mentions the trampoline being assembled from 00:45 to 01:15, but this is assembly, not use. The second result shows the trampoline in use from 02:30 to 02:55.  I will choose the second range as the answer.")
```
The first result mentions the trampoline being assembled from 00:36 to 01:59, but this is assembly, not use. The second result shows the trampoline in use from 02:30 to 02:55.  I will choose the second range as the answer.

```
final_answer = "The trampoline's first use is shown from 02:30 to 02:55. Final Answer: [02:30, 02:55]"
finish(final_answer)
```
The trampoline's first use is shown from 02:30 to 02:55. Final Answer: [02:30, 02:55]


Critique:
Strategies A and C actually perceive a person using the trampoline and pick the associated range. Strategy B only sees the finished trampoline and guesses based on that, but does not directly see the part relevant to the query.

Winning Strategies:
A, C


Question: When is the woman showing a harder variation of the exercise?
Video length: 01:18

Strategy A (get_segment, find_when):
```
input_video = get_segment(start='00:00', end='01:18')
```

```
harder_exercise_times = find_when(query='woman showing a harder variation of the exercise', video_segment=input_video)
```
["00:03", "00:27"], ["00:30", "01:15"]: The video shows a woman performing a modified push-up against a wall from 0:03 to 0:27. From 0:30 to 1:15, she performs standard push-ups on the floor.  The query asks for a harder version of the exercise. The standard push-up on the floor is considered more challenging than the modified version against the wall.

```
think("The tool `find_when` returned two time ranges: ['00:03', '00:27'] and ['00:30', '01:15']. The first range shows a modified push-up against a wall, and the second shows standard push-ups. The standard push-up is described as the harder variation. Therefore, the final answer should be the second range.")
```
The tool `find_when` returned two time ranges: ['00:03', '00:27'] and ['00:30', '01:15']. The first range shows a modified push-up against a wall, and the second shows standard push-ups. The standard push-up is described as the harder variation. Therefore, the final answer should be the second range.

```
final_answer = finish("The tool `find_when` returned two time ranges: ['00:03', '00:27'] and ['00:30', '01:15']. The first range shows a modified push-up against a wall, and the second shows standard push-ups. The standard push-up is described as the harder variation. Therefore, the final answer should be the second range. Final Answer: [00:30, 01:15]")
```
The tool `find_when` returned two time ranges: ['00:03', '00:27'] and ['00:30', '01:15']. The first range shows a modified push-up against a wall, and the second shows standard push-ups. The standard push-up is described as the harder variation. Therefore, the final answer should be the second range. Final Answer: [00:30, 01:15]

Strategy B (get_segment, find_when, retrieval_qa):
```
input_video = get_segment(start='00:00', end='01:18')
```


```
times = find_when(query='woman showing a harder variation of the exercise', video_segment=input_video)
```
["00:03", "00:27"], ["00:30", "01:15"]: The video shows a woman performing a modified push-up against a wall from 0:03 to 0:27. From 0:30 to 1:15, she performs standard push-ups on the floor.  The query asks for a harder version of the exercise. The standard push-up on the floor is considered more challenging than the modified version against the wall.

```
think("The tool `find_when` returned two time ranges: 00:03-00:27 and 00:30-01:15. The description indicates that 00:30-01:15 shows a harder variation of the exercise.  I will use this time range as the final answer.")
```
The tool `find_when` returned two time ranges: 00:03-00:27 and 00:30-01:15. The description indicates that 00:30-01:15 shows a harder variation of the exercise.  I will use this time range as the final answer.

```
final_answer = finish("The tool `find_when` returned two time ranges: 00:03-00:27 and 00:30-01:15. The description indicates that 00:30-01:15 shows a harder variation of the exercise.  I will use this time range as the final answer. Final Answer: [00:30, 01:15]")
```
The tool `find_when` returned two time ranges: 00:03-00:27 and 00:30-01:15. The description indicates that 00:30-01:15 shows a harder variation of the exercise.  I will use this time range as the final answer. Final Answer: [00:30, 01:15]

Strategy C (find_when):
```
input_video = get_segment(start='00:00', end='01:18')
```


```
think("I need to find when the woman is showing a harder variation of the exercise. I'll first try to find any mentions of 'harder' or similar terms related to the exercise difficulty.")
```
I need to find when the woman is showing a harder variation of the exercise. I'll first try to find any mentions of 'harder' or similar terms related to the exercise difficulty.

```
query_result = find_when(query="harder exercise", video_segment=None)
```
["00:03", "00:27"]: The query "harder exercise" is subjective. However, based on the video content, the frames from 00:03 to 00:27 show a person performing a modified exercise, which could be considered an "easier exercise." The exercise is shown for a sustained period.  Since a harder exercise was requested, it is likely shown at a different time in the video.

```
final_answer = finish("Based on the tool's analysis, the woman performs a harder version of the exercise between 00:03 and 00:27. The tool identified this segment, but it may have mistaken it for an easier variation. Final Answer: [00:03, 00:27]")
```
Based on the tool's analysis, the woman performs a harder version of the exercise between 00:03 and 00:27. The tool identified this segment, but it may have mistaken it for an easier variation. Final Answer: [00:03, 00:27]

Critique:
Strategies A and B find ranges corresponding to both versions of the exercise and compare them to find the harder variation.
Strategy C only identified one range, for a potentially easier version, then did not recognize the alternative to compare to.

Winning Strategies:
A, B

Question: When are we given a zoomed in view of the guitar strings?
Video length: 01:53

Strategy A (get_segment, find_when):
```
input_video = get_segment(start='00:00', end='01:53')
```


```
times = find_when(query='zoomed in view of the guitar strings', video_segment=input_video)
```
["00:00", "01:52"]: The video shows a person playing a guitar throughout its duration.  The query asks for a zoomed in view of the guitar strings. Therefore, the entire video shows the event.

```
final_answer = finish("The query asks for when a zoomed in view of the guitar strings is given. The response from `find_when` indicates that a guitar is played throughout the entire video. Therefore, the entire video is relevant. Final Answer: [00:00, 01:53]")
```
The query asks for when a zoomed in view of the guitar strings is given. The response from `find_when` indicates that a guitar is played throughout the entire video. Therefore, the entire video is relevant. Final Answer: [00:00, 01:53]

Strategy B (get_segment, find_when, retrieval_qa):
```
input_video = get_segment(start='00:00', end='01:53')
```


```
times = find_when(query='zoomed in view of the guitar strings', video_segment=input_video)
```
["00:00", "01:52"]: The video shows a person playing a guitar throughout its duration.  The query asks for a zoomed in view of the guitar strings. Therefore, the entire video shows the event.

```
final_answer = finish(final_answer="The video shows a person playing a guitar throughout its duration. The query asks for a zoomed in view of the guitar strings. Therefore, the entire video shows the event. Final Answer: [00:00, 01:53]")
```
The video shows a person playing a guitar throughout its duration. The query asks for a zoomed in view of the guitar strings. Therefore, the entire video shows the event. Final Answer: [00:00, 01:53]

Strategy C (find_when):
```
input_video = get_segment(start='00:00', end='01:53')
```

```
think("I need to find when the video shows a close-up of the guitar strings. I'll use the `find_when` tool to search for this.")
```
I need to find when the video shows a close-up of the guitar strings. I'll use the `find_when` tool to search for this.

```
find_when(query="close-up of guitar strings", video_segment=None)
```
["01:10", "01:25"]:  The video shows a person playing guitar. From [01:10] to [01:25], the camera zooms in to show a close-up view of the guitar strings as the person plays them.


```
think("The `find_when` tool returned a single time range: 01:10-01:25.  This seems to directly answer the question. I will use this as my final answer.")
```
The `find_when` tool returned a single time range: 01:10-01:25.  This seems to directly answer the question. I will use this as my final answer.

```
finish("The find_when tool identified the time range 01:10-01:25 as showing a close-up of the guitar strings. Final Answer: [01:10, 01:25]")
```
The find_when tool identified the time range 01:10-01:25 as showing a close-up of the guitar strings. Final Answer: [01:10, 01:25]

Critique:
Strategies A and B choose the whole video, despite the query asking for a zoomed in view, implying a relative change in the video.
Strategy C instead identifies a particular range.

Winning Strategies:
C

Input:
Question: When is the person least energetic in the video?
Video length: 00:49

Strategy A (get_segment, find_when):
```
input_video = get_segment(start='00:00', end='00:49')
```


```
activity_times = find_when(query='person least energetic', video_segment=input_video)
```
["00:11", "00:18"], ["00:19", "00:24"], ["00:33", "00:41"]: The query asks to identify the time ranges where the person is least energetic. Based on the provided video frames, the person appears to be sleeping during the periods from 00:11 to 00:18, from 00:19 to 00:24, and from 00:33 to 00:41. These segments represent periods of minimal movement and apparent rest.

```
think("The tool returned multiple time ranges: ['00:11', '00:18'], ['00:19', '00:24'], ['00:33', '00:41'].  Since the question asks for the time the person is LEAST energetic, and the first two ranges are contiguous, I will combine them into a single range.")
```
The tool returned multiple time ranges: ['00:11', '00:18'], ['00:19', '00:24'], ['00:33', '00:41'].  Since the question asks for the time the person is LEAST energetic, and the first two ranges are contiguous, I will combine them into a single range.

```
final_answer = "The person is least energetic from 00:11 to 00:24, and from 00:33 to 00:41. Final Answer: [00:11, 00:24], [00:33, 00:41]"
finish(final_answer)
```
The person is least energetic from 00:11 to 00:24, and from 00:33 to 00:41. Final Answer: [00:11, 00:24], [00:33, 00:41]

Strategy B (get_segment, find_when, retrieval_qa):
```
input_video = get_segment(start='00:00', end='00:49')
```


```
activity_times = find_when(query='person is least energetic', video_segment=input_video)
```
["00:11", "00:18"]: The query asks to identify when a person is least energetic in the video. The video shows a person who appears to be sleeping between 00:11 and 00:18.  This is the period of least visible movement and likely the time of lowest energy.

```
think("The find_when tool suggests the person is least energetic between 00:11 and 00:18.  This seems reasonable based on the description. I will use this as my final answer.")
```
The find_when tool suggests the person is least energetic between 00:11 and 00:18.  This seems reasonable based on the description. I will use this as my final answer.

```
final_answer = finish("Final Answer: [00:11, 00:18]")
```
Final Answer: [00:11, 00:18]

Strategy C (find_when):
```
input_video = get_segment(start='00:00', end='00:49')
```


```
think("I need to find when the person is least energetic. I'll use find_when to search for inactivity.")
```
I need to find when the person is least energetic. I'll use find_when to search for inactivity.

```
activity_times = find_when(query="person is least energetic", video_segment=None)
```
["00:11", "00:18"]: The query asks to identify when a person is least energetic in the video. The person appears to be sleeping between 00:11 and 00:18, indicating low energy. There may be other periods of low activity, but this period of apparent sleep is the clearest indication of low energy.

```
think("The find_when tool suggests the person is least energetic between 00:11 and 00:18. This seems like the most obvious period of inactivity (sleeping). I will use this as my final answer.")
```
The find_when tool suggests the person is least energetic between 00:11 and 00:18. This seems like the most obvious period of inactivity (sleeping). I will use this as my final answer.

```
final_answer = finish("The find_when tool suggests the person is least energetic between 00:11 and 00:18. This seems like the most obvious period of inactivity (sleeping). Therefore, I'm choosing this as the final answer. Final Answer: [00:11, 00:18]")
```
The find_when tool suggests the person is least energetic between 00:11 and 00:18. This seems like the most obvious period of inactivity (sleeping). Therefore, I'm choosing this as the final answer. Final Answer: [00:11, 00:18]

Winning Strategies:
A
\end{lstlisting}

\lstset{
        basicstyle=\ttfamily\footnotesize,
        keywordstyle=\ttfamily,
        commentstyle=\ttfamily,
        stringstyle=\ttfamily,
        showstringspaces=false,
        frame=single,
        breaklines=true,
        captionpos=b,
        numbersep=5pt,
        keywords={}
        }
\begin{lstlisting}[caption=EgoSchema critic examples.,label=lst:egoschemacriticex]
Input:
You will be given a question about a video and five possible answer options, where C refers to the person wearing the camera. Question: What is the main action being performed by C in the video, and provide specific moments that support your observation.Possible answer choices:
(1) The main action performed by C is applying makeup. This is evident as C picks up a brush, looks in a mirror, and applies something to their face.
(2) C's main action in the video is eating a meal.  This is clearly evident because C lifts a fork to their mouth multiple times and chews.
(3) The main action performed by C in the video is reading a book. This is evident in the fact that C holds a book open, turns pages, and moves their eyes across the page.
(4) In the video, the main action performed by C is talking on the phone. This is evident because C holds a phone to their ear and speaks into it several times.
(5) The main action performed by C in the video is typing on a computer.  This is clearly seen as C sits at a desk, places their hands on a keyboard, and moves their fingers.
Video length: 01:48

Strategy A (retrieval_qa, get_segment):
```
input_video = get_segment(start='00:00', end='01:48')
```


```
answer = retrieval_qa(question='What is the main action being performed by C in the video, and provide specific moments that support your observation.', answer_options=['The main action performed by C is applying makeup. This is evident as C picks up a brush, looks in a mirror, and applies something to their face.', 'C\'s main action in the video is eating a meal.  This is clearly evident because C lifts a fork to their mouth multiple times and chews.', 'The main action performed by C in the video is reading a book. This is evident in the fact that C holds a book open, turns pages, and moves their eyes across the page.', 'In the video, the main action performed by C is talking on the phone. This is evident because C holds a phone to their ear and speaks into it several times.', 'The main action performed by C in the video is typing on a computer.  This is clearly seen as C sits at a desk, places their hands on a keyboard, and moves their fingers.'], video_segment=input_video)
```
C is clearly typing on a computer throughout the video.  They are sitting at a desk with their hands on a keyboard and their fingers moving. While there may be other actions present, the consistent and primary action is typing.

Final Answer: (5)

```
final_answer = finish(final_answer='C is clearly typing on a computer throughout the video.  They are sitting at a desk with their hands on a keyboard and their fingers moving. While there may be other actions present, the consistent and primary action is typing.\n\nFinal Answer: (5)')
```
C is clearly typing on a computer throughout the video.  They are sitting at a desk with their hands on a keyboard and their fingers moving. While there may be other actions present, the consistent and primary action is typing.

Final Answer: (5)

Strategy B (direct retrieval_qa):
C is clearly typing on a computer throughout the video. They are sitting at a desk with their hands on a keyboard and their fingers are moving.  This is the primary and most consistent action shown.


Final Answer: (5)

Strategy C (retrieval_qa, get_segment, find_when):
```
input_video = get_segment(start='00:00', end='01:48')
```


```
think("The question asks about the main action C performs. I'll use `find_when` to check for specific actions mentioned in the options, like 'applying makeup', 'eating', 'reading', 'talking on the phone', and 'typing'. This will help identify the main action.")
```
The question asks about the main action C performs. I'll use `find_when` to check for specific actions mentioned in the options, like 'applying makeup', 'eating', 'reading', 'talking on the phone', and 'typing'. This will help identify the main action.

```
makeup_times = find_when(query='C applies makeup', video_segment=input_video)
eating_times = find_when(query='C eats a meal', video_segment=input_video)
reading_times = find_when(query='C reads a book', video_segment=input_video)
phone_times = find_when(query='C talks on the phone', video_segment=input_video)
typing_times = find_when(query='C types on a computer', video_segment=input_video)
```
["00:12", "00:18"]: At around 00:12, C appears to pick up a small object, which is potentially a phone, and brings it to their ear. The interaction with the object continues until about 00:18.  Although it is not definitively clear whether C is talking, their actions suggest they may be on a phone call.

```
think("`find_when` only returned a possible phone call. The limited information isn't conclusive. Let's use `retrieval_qa` to check if C is focused on work, which could suggest typing.")
```
`find_when` only returned a possible phone call. The limited information isn't conclusive. Let's use `retrieval_qa` to check if C is focused on work, which could suggest typing.

```
work_focused = retrieval_qa(question="Is C focused on work in the video?", answer_options=["Yes", "No"], video_segment=input_video)
```
The video shows C sitting at a desk and using a computer. Although it's not shown exactly what they are doing, their posture and actions suggest they may be working.

Final Answer: (2)

```
think("The response suggests work at a desk and computer use. That supports typing.  However, the short phone interaction is also a possibility. Since 'eating' seems unlikely in a work context, I'll guess between phone and typing, leaning towards phone due to the `find_when` result.")
```
The response suggests work at a desk and computer use. That supports typing.  However, the short phone interaction is also a possibility. Since 'eating' seems unlikely in a work context, I'll guess between phone and typing, leaning towards phone due to the `find_when` result.

```
final_answer = finish(final_answer="The `find_when` result hinted at phone use. `retrieval_qa` suggested work at a desk with a computer, which supports 'typing.'  Lacking definitive proof, I'll guess 'talking on the phone' based on the slight `find_when` evidence.  Final Answer: (4)")
```
The `find_when` result hinted at phone use. `retrieval_qa` suggested work at a desk with a computer, which supports 'typing.'  Lacking definitive proof, I'll guess 'talking on the phone' based on the slight `find_when` evidence.  Final Answer: (4)

Critique:
A: 5. B: 5. C: 4. Strategy C attempts an overly complex strategy with the find_when tool and ends up guessing due to limited information.

Winning Strategies:
A, B

Input:
You will be given a question about a video and five possible answer options, where C refers to the person wearing the camera. Question: Describe how C prepared the ingredients and assembled the sandwich in the video, noting key steps and transitions.Possible answer choices:
(1) C prepared the ingredients by ordering them online and having them delivered. They then assembled the sandwich by layering the ingredients between two slices of bread.  They added condiments and cut the sandwich in half.
(2) C prepared the ingredients by taking pre-sliced ingredients from the refrigerator. They then assembled the sandwich by layering the ingredients between two slices of bread.  They added condiments and cut the sandwich in half.
(3) C prepared the ingredients by chopping vegetables and slicing meat. They then assembled the sandwich by layering the ingredients between two slices of bread.  They added condiments and cut the sandwich in half.
(4) C prepared the ingredients by microwaving pre-packaged meals. They then assembled the sandwich by layering the ingredients between two slices of bread.  They added condiments and cut the sandwich in half.
(5) C prepared the ingredients by going to a restaurant and getting takeout. They then assembled the sandwich by layering the ingredients between two slices of bread.  They added condiments and cut the sandwich in half.
Video length: 02:45

Strategy A (retrieval_qa, get_segment):
```
input_video = get_segment(start='00:00', end='02:45')
```


```
answer = retrieval_qa(question='Describe how C prepared the ingredients and assembled the sandwich in the video, noting key steps and transitions. ', answer_options=['C prepared the ingredients by ordering them online and having them delivered. They then assembled the sandwich by layering the ingredients between two slices of bread.  They added condiments and cut the sandwich in half.', 'C prepared the ingredients by taking pre-sliced ingredients from the refrigerator. They then assembled the sandwich by layering the ingredients between two slices of bread.  They added condiments and cut the sandwich in half.', 'C prepared the ingredients by chopping vegetables and slicing meat. They then assembled the sandwich by layering the ingredients between two slices of bread.  They added condiments and cut the sandwich in half.', 'C prepared the ingredients by microwaving pre-packaged meals. They then assembled the sandwich by layering the ingredients between two slices of bread.  They added condiments and cut the sandwich in half.', 'C prepared the ingredients by going to a restaurant and getting takeout. They then assembled the sandwich by layering the ingredients between two slices of bread.  They added condiments and cut the sandwich in half.'], video_segment=input_video)
```
The video shows C assembling a sandwich. They take various ingredients, like sliced meat and vegetables, from the refrigerator.  They layer these ingredients between two slices of bread, add condiments, and then cut the sandwich in half.  There is no indication of ordering food online, chopping vegetables, microwaving meals, or getting takeout.  The ingredients appear to be pre-sliced.

Final Answer: (2)

```
final_answer = finish("The video clearly shows C using pre-sliced ingredients from the refrigerator.  Therefore, the most accurate description is option (2). Final Answer: (2)")
```
The video clearly shows C using pre-sliced ingredients from the refrigerator.  Therefore, the most accurate description is option (2). Final Answer: (2)

Strategy B (direct retrieval_qa):
The video shows C making a sandwich.  They retrieve pre-sliced ingredients, like lettuce, tomato, and meat, from the refrigerator.  They then assemble the sandwich, add condiments, and cut it in half. The ingredients were prepared by being pre-sliced and stored in the refrigerator.

Final Answer: (3)

Strategy C (retrieval_qa, get_segment, find_when):
```
input_video = get_segment(start='00:00', end='02:45')
```


```
think("To understand how the ingredients were prepared, I'll use `find_when` to look for actions like chopping or retrieving from the refrigerator, and then use `retrieval_qa` to select the best answer.")
```
To understand how the ingredients were prepared, I'll use `find_when` to look for actions like chopping or retrieving from the refrigerator, and then use `retrieval_qa` to select the best answer.

```
preparation_times = find_when("C preparing the sandwich ingredients", input_video)
```
["00:05", "00:40"]: C retrieves pre-sliced sandwich ingredients from the refrigerator.

```
answer = retrieval_qa(question="Describe how C prepared the ingredients and assembled the sandwich in the video, noting key steps and transitions.", answer_options=["C prepared the ingredients by ordering them online and having them delivered. They then assembled the sandwich by layering the ingredients between two slices of bread.  They added condiments and cut the sandwich in half.", "C prepared the ingredients by taking pre-sliced ingredients from the refrigerator. They then assembled the sandwich by layering the ingredients between two slices of bread.  They added condiments and cut the sandwich in half.", "C prepared the ingredients by chopping vegetables and slicing meat. They then assembled the sandwich by layering the ingredients between two slices of bread.  They added condiments and cut the sandwich in half.", "C prepared the ingredients by microwaving pre-packaged meals. They then assembled the sandwich by layering the ingredients between two slices of bread.  They added condiments and cut the sandwich in half.", "C prepared the ingredients by going to a restaurant and getting takeout. They then assembled the sandwich by layering the ingredients between two slices of bread.  They added condiments and cut the sandwich in half."], video_segment=input_video)
```
The video shows C making a sandwich.  They retrieve pre-sliced ingredients, like lettuce, tomato, and meat, from the refrigerator.  They then assemble the sandwich, add condiments, and cut it in half. The ingredients were prepared by being pre-sliced and stored in the refrigerator.

Final Answer: (3)

```
final_answer = finish("Since C takes pre-sliced ingredients from the refrigerator, the correct answer is (3). Final Answer: (3)")
```
Since C takes pre-sliced ingredients from the refrigerator, the correct answer is (3). Final Answer: (3)

Critique:
A: 2. B: 3. C: 3. Strategy A correctly identifies that the ingredients are pre-sliced from the refrigerator but fails to choose the matching answer.

Winning Strategies:
B, C

Input:
You will be given a question about a video and five possible answer options. Question: Compare the lighting conditions in the room at the start of the video to those near the end. How has the lighting evolved over the course of the video? Possible answer choices:
(1) The lighting in the earlier segments of the video is primarily provided by a single overhead lamp, while the lighting in the latter segments is primarily provided by multiple spotlights.
(2) The lighting in the earlier segments of the video is primarily natural light from a window, while the lighting in the latter segments is primarily provided by an overhead lamp.
(3) The lighting in the earlier segments of the video is primarily provided by multiple spotlights, while the lighting in the latter segments is primarily natural light from a window.
(4) The lighting in the earlier segments of the video is primarily natural light from a window, while the lighting in the latter segments is dimmer, perhaps due to cloud cover or sunset.
(5) The lighting in the earlier segments of the video is dimmer, perhaps due to cloud cover or sunset, while the lighting in the latter segments is primarily natural light from a window.
Video length: 03:01

Strategy A (retrieval_qa, get_segment):
```
input_video = get_segment(start='00:00', end='03:01')
```


```
think("To compare the lighting in the earlier vs. latter segments, I'll first use retrieval_qa on the first minute and the last minute.  This should give a general idea of the differences. Then I can refine my approach if needed.")
```
To compare the lighting in the earlier vs. latter segments, I'll first use retrieval_qa on the first minute and the last minute.  This should give a general idea of the differences. Then I can refine my approach if needed.

```
first_minute_segment = get_segment(start='00:00', end='01:00')
first_minute_answer = retrieval_qa(question="Describe the lighting in this segment.", video_segment=first_minute_segment)

last_minute_segment = get_segment(start='02:00', end='03:00')
last_minute_answer = retrieval_qa(question="Describe the lighting in this segment.", video_segment=last_minute_segment)

think(f"First minute: {{first_minute_answer}}. Last minute: {{last_minute_answer}}. Now I need to compare these and select the best matching answer choice.")
```
First minute: The lighting appears to be dim, likely from a weak overhead light source.. Last minute: The room is brightly lit by natural light coming in from a large window.. Now I need to compare these and select the best matching answer choice.

```
think("The first minute shows dim lighting, which could be interpreted as cloud cover or sunset. The last minute shows bright natural light.  Option (4) seems close, mentioning dimmer lighting initially and then a change, but it doesn't specifically mention the natural light. However, it's the closest option.")
```
The first minute shows dim lighting, which could be interpreted as cloud cover or sunset. The last minute shows bright natural light.  Option (4) seems close, mentioning dimmer lighting initially and then a change, but it doesn't specifically mention the natural light. However, it's the closest option.

```
final_answer = finish("Reasoning: The initial dim lighting suggests cloud cover or sunset as per option (4), even though the later segment has bright natural light, not just a generic change. Final Answer: (4)")
```
Reasoning: The initial dim lighting suggests cloud cover or sunset as per option (4), even though the later segment has bright natural light, not just a generic change. Final Answer: (4)


Strategy B (direct retrieval_qa):
The video starts with dim lighting, likely due to cloud cover or the time of day. As the video progresses, the lighting becomes brighter and it appears more natural light is entering the room, likely from a window. This suggests the sun might have come out from behind the clouds or the video was filmed as the day progressed.


Final Answer: (5)


Strategy C (retrieval_qa, get_segment, find_when):
```
input_video = get_segment(start='00:00', end='03:01')
```


```
think("To compare the lighting, I'll analyze the first and last minutes using find_when and then use retrieval_qa for a more detailed comparison.")
```
To compare the lighting, I'll analyze the first and last minutes using find_when and then use retrieval_qa for a more detailed comparison.

```
early_segment = get_segment(start='00:00', end='01:00')
```
['00:00', '01:00']

```
early_lighting = find_when(query="Describe the lighting conditions.", video_segment=early_segment)
```
["00:00", "00:59"]: The lighting conditions appear dim, likely due to a weak overhead light source or cloud cover outside.

```
late_segment = get_segment(start='02:00', end='03:01')
```
['02:00', '03:01']

```
late_lighting = find_when(query="Describe the lighting conditions.", video_segment=late_segment)
```
["02:00", "03:00"]: The room is brightly lit by natural light streaming in from a large window.


```
comparison = retrieval_qa(question="Compare the lighting conditions in the room at the start of the video to those near the end. How has the lighting evolved over the course of the video?", answer_options=[
    "The lighting in the earlier segments of the video is primarily provided by a single overhead lamp, while the lighting in the latter segments is primarily provided by multiple spotlights.",
    "The lighting in the earlier segments of the video is primarily natural light from a window, while the lighting in the latter segments is primarily provided by an overhead lamp.",
    "The lighting in the earlier segments of the video is primarily provided by multiple spotlights, while the lighting in the latter segments is primarily natural light from a window.",
    "The lighting in the earlier segments of the video is primarily natural light from a window, while the lighting in the latter segments is dimmer, perhaps due to cloud cover or sunset.",
    "The lighting in the earlier segments of the video is dimmer, perhaps due to cloud cover or sunset, while the lighting in the latter segments is primarily natural light from a window."
], video_segment=input_video)
```
The video starts with dim lighting, likely due to cloud cover or the time of day. As the video progresses, the lighting becomes brighter and it appears more natural light is entering the room, likely from a window. This suggests the sun might have come out from behind the clouds or the video was filmed as the day progressed.

Final Answer: (5)

```
final_answer = finish("Based on the analysis, the early segment has dim lighting (potentially due to cloud cover), while the late segment is lit by natural light. This matches option (5). Final Answer: (5)")
```
Based on the analysis, the early segment has dim lighting (potentially due to cloud cover), while the late segment is lit by natural light. This matches option (5). Final Answer: (5)


Critique: A: 4. B: 5. C: 5. Strategy A initially uses the valid strategy of using segments at the beginning and end, but ultimately does not logically connect the information to the appropriate answer choice.

Winning Strategies: B, C 

Input:
You will be given a question about a video and five possible answer options, where C refers to the person wearing the camera. Question: What is the primary action performed with the backpack at the beginning and the main purpose of using a backpack in the context of this video? please highlight the significance of this action in the overall travel process.Possible answer choices:
(1) The primary action performed with the backpack at the beginning is to open it. the main purpose of using a backpack in the context of this video is to store belongings. opening the backpack allows for easy access to items needed for the trip.
(2) The primary action performed with the backpack at the beginning is to close it. the main purpose of using a backpack in the context of this video is to carry belongings. closing the backpack ensures that items are secure and do not fall out during travel.
(3) The primary action performed with the backpack at the beginning is to put it on. the main purpose of using a backpack in the context of this video is to carry belongings. putting the backpack on allows for hands-free travel and even weight distribution.
(4) The primary action performed with the backpack at the beginning is to pack it. the main purpose of using a backpack in the context of this video is to carry belongings. packing the backpack ensures that all necessary items are brought on the trip.
(5) The primary action performed with the backpack at the beginning is to take it off. the main purpose of using a backpack in the context of this video is to carry belongings. taking the backpack off allows for a rest and reduces strain on the back.
Video length: 02:48

Strategy A (retrieval_qa, get_segment):
```
input_video = get_segment(start='00:00', end='02:48')
```


```
answer = retrieval_qa(question='What is the primary action performed with the backpack at the beginning and the main purpose of using a backpack in the context of this video? please highlight the significance of this action in the overall travel process.', answer_options=['The primary action performed with the backpack at the beginning is to open it. the main purpose of using a backpack in the context of this video is to store belongings. opening the backpack allows for easy access to items needed for the trip.', 'The primary action performed with the backpack at the beginning is to close it. the main purpose of using a backpack in the context of this video is to carry belongings. closing the backpack ensures that items are secure and do not fall out during travel.', 'The primary action performed with the backpack at the beginning is to put it on. the main purpose of using a backpack in the context of this video is to carry belongings. putting the backpack on allows for hands-free travel and even weight distribution.', 'The primary action performed with the backpack at the beginning is to pack it. the main purpose of using a backpack in the context of this video is to carry belongings. packing the backpack ensures that all necessary items are brought on the trip.', 'The primary action performed with the backpack at the beginning is to take it off. the main purpose of using a backpack in the context of this video is to carry belongings. taking the backpack off allows for a rest and reduces strain on the back.'], video_segment=input_video)
```
The video shows a person packing a backpack at the beginning. They carefully organize and place items inside. This is essential for travel as it ensures all needed items are brought along. The main purpose of the backpack in the video is to carry belongings.

Final Answer: (4)

```
think("The response aligns with option 4, showing the backpack is initially packed, crucial for travel to ensure having necessary items. The main purpose is indeed carrying belongings.")
```
The response aligns with option 4, showing the backpack is initially packed, crucial for travel to ensure having necessary items. The main purpose is indeed carrying belongings.

```
final_answer = finish("Final Answer: (4)")
```
Final Answer: (4)

Strategy B (direct retrieval_qa):
The video shows a person packing a backpack at the beginning. They carefully organize and place items inside. This is essential for travel as it ensures all needed items are brought along. The main purpose of the backpack in the video is to carry belongings.


Final Answer: (4)

Strategy C (retrieval_qa, get_segment, find_when):
```
input_video = get_segment(start='00:00', end='02:48')
```


```
think("I'll use `find_when` to identify the initial backpack action and `retrieval_qa` to determine its purpose in the video, then analyze the options.")
```
I'll use `find_when` to identify the initial backpack action and `retrieval_qa` to determine its purpose in the video, then analyze the options.

```
initial_action_times = find_when("someone interacts with a backpack", input_video)
```
["00:05", "00:25"]:  The video shows a person actively packing a backpack. They are placing clothes, toiletries, and other items inside the backpack.

["01:50", "01:55"]: The person is seen wearing the backpack and walking through an airport terminal.


```
purpose = retrieval_qa(question="What is the main purpose of using a backpack in the context of this video?", answer_options=[
    "The main purpose of using a backpack in the context of this video is to store documents.",
    "The main purpose of using a backpack in the context of this video is to carry belongings.",
    "The main purpose of using a backpack in the context of this video is to organize toiletries.",
    "The main purpose of using a backpack in the context of this video is to conceal valuables.",
    "The main purpose of using a backpack in the context of this video is to transport food."
], video_segment=input_video)
```
The video shows a person packing and then carrying a backpack filled with personal belongings during their travel. Therefore, the main purpose of the backpack is to carry belongings.

Final Answer: (2)

```
final_answer = finish("`find_when` shows the backpack is first packed, and `retrieval_qa` confirms it's used to carry belongings.  Knowing packing is essential for bringing necessary items, option (3), putting on the backpack for hands-free carrying, aligns with the main purpose and initial action. Final Answer: (3)")
```
`find_when` shows the backpack is first packed, and `retrieval_qa` confirms it's used to carry belongings.  Knowing packing is essential for bringing necessary items, option (3), putting on the backpack for hands-free carrying, aligns with the main purpose and initial action. Final Answer: (3)

Critique:
A: 4. B: 4. C: 2. Strategy C correctly finds that the backpack is first packed and then used to carry belongings, and reasons that the main purpose is to carry belongings, but ultimately illogically concludes the purpose is to put the backpack on despite this.

Winning Strategies:
A, B
""").strip()

  else:
    CRITIC_EXAMPLES = textwrap.dedent(f"""
  Input:
  You will be given a question about a video and four possible answer options. Question: At 12:10, what color suit is the host wearing?Possible answer choices:
  (1) Navy
  (2) Tan
  (3) Black
  (4) White
  Video length: 40:50

  Strategy A (retrieval_qa, get_segment):
  ```
  input_video = get_segment(start='00:00', end='40:50')
  ```


  ```
  segment = get_segment(start='12:10', end='12:20')
  answer = retrieval_qa(question='What color suit is the host wearing?', video_segment=segment)
  ```
  The host is wearing a navy suit.

  ```
  final_answer = finish(f"The retrieval_qa tool determined the host was wearing a navy suit at 12:10. Final Answer: (1)")
  ```
  The retrieval_qa tool determined the host was wearing a navy suit at 12:10. Final Answer: (1)

  Strategy B (direct retrieval_qa):
  Here's how to solve this:

  1. **Locate the timestamp:** Go to the 12:10 mark in the video.

  2. **Observe the host's attire:** At this point, the host (Jimmy Kimmel) is wearing a dark suit and a dark-colored tie.

  3. **Compare to options:**  The suit appears to be black.

  Therefore, the correct answer is (3) Black.
  Final Answer: (3)

  Strategy C (retrieval_qa, get_segment, find_when):
  ```
  input_video = get_segment(start='00:00', end='40:50')
  ```


  ```
  segment = get_segment(start='12:10', end='12:20')
  ```
  ['12:10', '12:20']

  ```
  answer = retrieval_qa(question='At 12:10, what color suit is the host wearing?', answer_options=['Navy', 'Tan', 'Black', 'White'], video_segment=segment)
  ```
  The image at 12:10 shows the host wearing a navy suit.

  Final Answer: (1)

  ```
  final_answer = finish(final_answer='The image at 12:10 shows the host wearing a navy suit. Final Answer: (1)')
  ```
  The image at 12:10 shows the host wearing a navy suit. Final Answer: (1)

  Critique:
  A: 1. B: 3. C: 1. Strategies A and C succeed while Strategy B fails because Strategies A and C successfully use the get_segment tool to isolate the relevant part of the video.
  Strategy B fails as the retrieval_qa tool alone cannot actually locate a given timestamp in a video, resulting in a hallucinated answer.

  Winning Strategies:
  A, C

  Input:
  You will be given a question about a video and four possible answer options. Question: What does the man in charcoal suit do in the next scene after the scene of a photo of six people end?Possible answer choices:
  (1) He shows his scars
  (2) He takes off his shirt
  (3) He shows his jewelry
  (4) He shows his physique
  Video length: 39:59

  Strategy A (retrieval_qa, get_segment):
  ```
  input_video = get_segment(start='00:00', end='39:59')
  ```


  ```
  think("I need to identify the scene where a photo of six people is shown.  Then I need to find the scene immediately following it to see what the man in the charcoal suit does.")
  ```
  I need to identify the scene where a photo of six people is shown.  Then I need to find the scene immediately following it to see what the man in the charcoal suit does.

  ```
  segment_with_photo = get_segment(start='00:00', end='01:00')
  answer = retrieval_qa(question="What does the man in charcoal suit do in the scene immediately after the scene with a photo of six people?", video_segment=segment_with_photo)
  ```
  After the scene with the photo of six people, the man in the charcoal suit walks down a hallway and is greeted by a group of people who clap and cheer for him.

  ```
  think("The provided text doesn't describe any of the options (1) He shows his scars, (2) He takes off his shirt, (3) He shows his jewelry, (4) He shows his physique.  The man is celebrated.  I will guess the closest option, which is none of them, but I will choose the least violent option.")
  ```
  The provided text doesn't describe any of the options (1) He shows his scars, (2) He takes off his shirt, (3) He shows his jewelry, (4) He shows his physique.  The man is celebrated.  I will guess the closest option, which is none of them, but I will choose the least violent option.

  ```
  final_answer = finish("Final Answer: (1)")
  ```
  Final Answer: (1)

  Strategy B (direct retrieval_qa):
  Here's how to solve this step-by-step:

  1. **Identify the relevant scene:** The question refers to the scene after a photo of six people.  Looking through the timestamps, this appears to be around 20:49.

  2. **Observe the next scene:** The next scene shows the man in the charcoal suit (Harry Styles) pointing to and showing jewelry on his arm.

  3. **Match to the answer choices:** This action directly corresponds to answer choice (3).

  Therefore, the correct answer is **(3) He shows his jewelry**.
  Final Answer: (3)

  Strategy C (retrieval_qa, get_segment, find_when):
  ```
  input_video = get_segment(start='00:00', end='39:59')
  ```


  ```
  think("First, I need to find when the scene with the photo of six people ends.  Then I can examine the next scene to see what the man in the charcoal suit does.")
  ```
  First, I need to find when the scene with the photo of six people ends.  Then I can examine the next scene to see what the man in the charcoal suit does.

  ```
  photo_end_time = find_when(query="scene with a photo of six people", video_segment=None)
  ```
  ["00:05", "00:24"]: The query asks to find the scene with a photo of six people.  From the provided frames, a photo of six people is visible on the wall in the background starting at 00:05 and remaining visible until 00:24.  The photo is partially obscured at times by other people and objects, but it is consistently present in the background during this timeframe.
  ["10:39", "10:43"]: The query asks for a scene with a photo of six people.  At 10:39 through 10:43 a large group photo of more than six people is shown.
  ["17:47", "17:53"]: The query asks for a scene with a photo of six people.  At 17:47-17:53, a photo of approximately six people is visible behind a curtain as Harry Styles walks out onto the stage.  The photo is partially obscured and the exact number of people is difficult to ascertain with certainty, but it appears to show at least five.
  ["18:47", "19:59"]: The query asks for the time range of a scene with a photo of six people.  A photo of what appears to be six people is visible on the wall behind the main set in frames starting at 18:47 and continuing through 19:59.  The photo is partially obscured at times by other elements in the scene, but it is consistently visible enough to be identified.
  ["20:03", "20:04"]: The query asks for the time range of a scene with a photo of six people.  At 20:03, a wide shot of the studio audience is shown, and a photo of six people is visible on a screen in the background. This photo remains visible until approximately 20:04.
  ["23:15", "23:25"]: The query asks for the timestamp of a scene with a photo of six people.  In the provided video frames, a photo of six people is visible on the wall in the background starting at 23:15 and remaining visible until 23:25.  The photo is partially obscured at times by other elements in the frame, but it is consistently present during this time range.
  ["35:17", "35:20"]: The provided video frames show a wide shot of a studio audience at the 35:17 mark.  The audience appears to consist of more than six people. While the exact number is difficult to determine from the provided frames, it is clearly a large group. Therefore, this time range satisfies the query.
  ["39:03", "39:18"]: The provided video frames show a live performance with an audience.  At the end of the performance, from 39:03 to 39:18, the camera focuses on the audience, and there are enough people visible to potentially include a group of five.  The resolution isn't high enough to definitively confirm the presence of a specific photo of six people, but the scene depicts a large group of people, making it a plausible match for the query.

  ```
  think("The most likely candidate for the 'scene with a photo of six people' is the longest continuous segment: 18:47-19:59. Let's use that as the end point for the previous scene.")
  ```
  The most likely candidate for the 'scene with a photo of six people' is the longest continuous segment: 18:47-19:59. Let's use that as the end point for the previous scene.

  ```
  next_scene_segment = get_segment(start='19:59', end='20:05')
  ```
  ['19:59', '20:05']

  ```
  answer = retrieval_qa(question="What does the man in the charcoal suit do in the scene from 19:59 to 20:05?", answer_options=['He shows his scars', 'He takes off his shirt', 'He shows his jewelry', 'He shows his physique'], video_segment=next_scene_segment)
  ```
  The provided video frames show a talk show segment featuring three men.  From 19:59 to 20:02, the man in the gray suit (Harry Styles) sits on a couch and gestures with his hands.  At 20:03, the camera angle changes to show the audience, and the three men are still on the couch.  There is no indication that the man in the gray suit shows his scars, takes off his shirt, shows a tattoo, or shows his muscles.  He is simply participating in a conversation.

  Final Answer: (4)

  ```
  final_answer = finish("Based on the analysis of the video segment from 19:59 to 20:05, the man in the charcoal suit (assuming this refers to one of the men in the talk show segment, as there's no charcoal suit explicitly mentioned) does not perform any of the actions listed in the options. However, since an answer must be chosen, and the retrieval_qa tool did not provide a definitive answer, I will select the option that is most likely to be present in a talk show setting, which is showing his physique.  This is a weak inference, but given the constraints, it's the best guess. Final Answer: (4)")
  ```
  Based on the analysis of the video segment from 19:59 to 20:05, the man in the charcoal suit (assuming this refers to one of the men in the talk show segment, as there's no charcoal suit explicitly mentioned) does not perform any of the actions listed in the options. However, since an answer must be chosen, and the retrieval_qa tool did not provide a definitive answer, I will select the option that is most likely to be present in a talk show setting, which is showing his physique.  This is a weak inference, but given the constraints, it's the best guess. Final Answer: (4)

  Critique:
  A:1. B: 3. C: 4. Strategy B succeeds while Strategies A fails because Strategy A incorrectly applies the get_segment tool to the first minute of the video and only considers that, resulting in it not being able to find a correct answer and ultimately guessing. Strategy C applies the find_when tool, then trims a segment and does not find the relevant information, then guessing as well.

  Winning Strategies:
  B

  Input:
  You will be given a question about a video and four possible answer options. Question: What is the main reason the vlogger goes out again, where she also buys makeup?Possible answer choices:
  (1) To go exercise
  (2) To eat chicken
  (3) To buy coffee mugs
  (4) To go to the park
  Video length: 30:25

  Strategy A (retrieval_qa, get_segment):
  ```
  input_video = get_segment(start='00:00', end='30:25')
  ```


  ```
  think("I need to find a segment of the video where the vlogger buys makeup.  I'll try to identify this segment using retrieval_qa, and then I can use that information to answer the question about why she went out.")
  ```
  I need to find a segment of the video where the vlogger buys makeup.  I'll try to identify this segment using retrieval_qa, and then I can use that information to answer the question about why she went out.

  ```
  answer = retrieval_qa(question="When does the vlogger buy makeup?", answer_options=None, video_segment=input_video)
  ```
  The vlogger buys makeup at 16:10-16:11 and 17:20-17:25.

  ```
  think("The vlogger buys makeup in two segments: 16:10-16:11 and 17:20-17:25.  I'll use retrieval_qa again, focusing on these segments, to determine the reason for her outing.")
  ```
  The vlogger buys makeup in two segments: 16:10-16:11 and 17:20-17:25.  I'll use retrieval_qa again, focusing on these segments, to determine the reason for her outing.

  ```
  segment1 = get_segment(start='16:10', end='16:11')
  segment2 = get_segment(start='17:20', end='17:25')
  reason = retrieval_qa(question="What is the main reason the vlogger goes out again, where she also buys makeup?", answer_options=['To go exercise', 'To eat chicken', 'To buy coffee mugs', 'To go to the park'], video_segment=segment1)
  if reason == "": #check if the first segment didn't provide enough information
      reason = retrieval_qa(question="What is the main reason the vlogger goes out again, where she also buys makeup?", answer_options=['To go exercise', 'To eat chicken', 'To buy coffee mugs', 'To go to the park'], video_segment=segment2)
  ```
  None

  ```
  final_answer = finish("Based on the analysis of the video segments where the vlogger buys makeup, I was unable to definitively determine the main reason for her outing from the provided options. However, given the context of beauty product purchases, it's less likely she's going to the gym, eating salmon, or going to the park. Therefore, I will eliminate those options.  Final Answer: (3)")
  ```
  Based on the analysis of the video segments where the vlogger buys makeup, I was unable to definitively determine the main reason for her outing from the provided options. However, given the context of beauty product purchases, it's less likely she's going to the gym, eating salmon, or going to the park. Therefore, I will eliminate those options.  Final Answer: (3)

  Strategy B (direct retrieval_qa):
  Here's how to solve this step-by-step:

  1. **Identify the relevant section:** The question asks about the vlogger's second outing, where she buys makeup.  This occurs after the gym visit.

  2. **Review the frames:** The frames show the vlogger going to the park after the gym.  She then goes shopping and buys makeup at Kicks.

  3. **Select the answer:** The main reason for her second outing is to go to the park.

  Therefore, the correct answer is (4) To go to the park.
  Final Answer: (4)

  Strategy C (retrieval_qa, get_segment, find_when):
  ```
  input_video = get_segment(start='00:00', end='30:25')
  ```


  ```
  think("I need to find when the vlogger goes out and buys makeup.  I'll use `find_when` to locate that event, then use `retrieval_qa` to determine the reason.")
  ```
  I need to find when the vlogger goes out and buys makeup.  I'll use `find_when` to locate that event, then use `retrieval_qa` to determine the reason.

  ```
  when_out_buying_makeup = find_when(query="vlogger goes out and buys makeup", video_segment=input_video)
  ```
  ["00:01", "01:59"]: The video shows a vlogger waking up in her room, doing some laundry, and then seemingly getting ready to go out.  While there's no explicit depiction of buying makeup, the context strongly suggests this is part of her preparation. The time range encompasses her actions from waking up to preparing laundry, which could be interpreted as part of her routine before going out to buy makeup.
  ["02:19", "02:25"]: The video shows the vlogger in her bathroom at 02:15, where she picks up a toothbrush and looks at her tablet.  From 02:19-02:25, she is shown selecting items from a bathroom cabinet. This is consistent with the query.
  ["05:24", "05:31"]: The vlogger is seen leaving her apartment at 04:02 and returning at 04:48.  At 05:24-05:31, she is seen putting away what appears to be a package of makeup on a shelf in her bathroom. This is supported by the context of the query and the visual of a package being placed on a shelf.
  ["08:51", "09:16"]: The video shows a vlogger leaving their home at 08:30, walking outside, and then entering a grocery store at 08:51.  The vlogger is seen selecting various grocery items, including celery, grapes, blueberries, carrots, broccoli, and lettuce, between 08:51 and 09:16.  Therefore, the time range reflects the vlogger's grocery shopping trip, which aligns with the query's description of buying products.
  ["14:00", "14:02"]: The query asks about the vlogger buying makeup.  The provided video frames show the vlogger in their home, interacting with items that could be interpreted as makeup (frames 14:00-14:02) and later organizing clothes in their closet (frames 14:39-15:25). There is no visual evidence of the vlogger going out to buy anything. Therefore, only the initial interaction with what appears to be makeup is considered relevant.
  ["16:09", "16:18"], ["17:17", "17:25"]: The video shows the vlogger going to a store, Kicks, and buying makeup.  The vlogger is seen in the store from 16:09 to 16:18. At 17:17-17:25, the vlogger unpacks the Kicks bag, revealing makeup.
  ["18:12", "18:17"]: The query asks for the time range when the vlogger goes out and buys makeup.  The provided video frames show the vlogger in their room, at their vanity. At 18:12, there are shopping bags from the store Kicks on the vanity. This strongly suggests that the vlogger has recently returned from a shopping trip where they purchased makeup from Kicks.  There is no footage of the actual shopping trip itself within this video chunk.
  ["20:25", "20:51"]: The video shows a person preparing coffee.  From 20:25 to 20:51, the person is shown unpacking two ceramic brown mugs. The contents of the mugs are visible and appear to be coffee.  Therefore, this time range is identified as the time when the vlogger is shown preparing coffee.
  ["29:20", "29:41"]: The video shows the vlogger in her bathroom at 28:00-28:18 applying makeup and then brushing her teeth.  At 29:20-29:41, she is shown painting her nails.  The presence of various makeup in both locations supports the query.

  ```
  reason = retrieval_qa(question="What is the main reason the vlogger goes out again, where she also buys makeup?", answer_options=['To go exercise', 'To eat chicken', 'To buy coffee mugs', 'To go to the park'], video_segment=get_segment(start='16:09', end='17:25'))
  ```
  The vlogger goes to a store and buys coffee mugs. Then she goes to another store and buys makeup. Therefore, the main reason she goes out again is to buy coffee mugs.

  Final Answer: (3)

  ```
  final_answer = finish("The provided text from `retrieval_qa` indicates the vlogger buys makeup at a store called Kicks.  However, it also mentions buying coffee mugs. The question asks for the *main* reason for going out again. While makeup is mentioned, the provided text strongly suggests the primary reason for the second outing is to buy coffee mugs. Therefore, the answer is (3). Final Answer: (3)")
  ```
  The provided text from `retrieval_qa` indicates the vlogger buys makeup at a store called Kicks.  However, it also mentions buying coffee mugs. The question asks for the *main* reason for going out again. While makeup is mentioned, the provided text strongly suggests the primary reason for the second outing is to buy coffee mugs. Therefore, the answer is (3). Final Answer: (3)

  Critique:
  A: 3. B: 4. C: 3.
  Strategy B succeeds while Strategies A fails because Strategy A looks at particular time ranges of the video and concludes it is unable to find a correct answer, thus guessing.
  While the approach is generally sound, Strategy A gives up upon not finding an answer initially. Strategy C looks at a narrow range and follows an implication to guess an answer. Thus Strategy B is preferred.

  Winning Strategy:
  B
\end{lstlisting}

\section{Additional Results}

\subsection{EgoSchema}

We also report results on the EgoSchema dataset \cite{mangalam2023egoschema} for comparison with other approaches.

\begin{table}
    \caption{\textbf{EgoSchema Results}. We report accuracy on the evaluation subset of $500$. Direct inference results use Gemini 1.5 Flash. Other results reported from respective method papers. }%
    \centering
\begin{tabular}{c l c}
        \toprule
        & &\textbf{Accuracy (\%)} $\uparrow$ \\
        \midrule
        & SeViLA \cite{yu2023self} & 25.7 \\
        & LLoVi \cite{zhang2023simple} & 57.6 \\
        & VideoAgent \cite{wang2024videoagent} & 60.2 \\
        & VideoTree \cite{wang2024videotree} & 66.2 \\
        \midrule
        & Direct Inference & 70.6 \\
        & \methodname & \textbf{72.2} \\
        \bottomrule
    \end{tabular}
    \label{tab:egoschema}
\end{table}

\subsection{GPT-4o-mini}

To show the capabilities of our method with other base models, we obtained preliminary results with the GPT-4o-mini model. Due to cost, lack of credit availability, and query limits, we perform this proof of concept experiment on 25\% of LVBench. We see a substantial gain in performance.

\begin{table}
    \caption{\textbf{LVBench GPT-4o-mini Results}. We report accuracy on 25\% of the evaluation set. Direct inference results use GPT-4o-mini. Other results reported from respective method papers. }%
    \centering\begin{tabular}{c l c}
        \toprule
        & &\textbf{Accuracy (\%)} $\uparrow$ \\
        \midrule
        & Direct Inference & 43.2 \\
        & \methodname & \textbf{49.0} \\
        \bottomrule
    \end{tabular}
    \label{tab:4omini}
\end{table}

\bibliography{11_references.bib}
\bibliographystyle{plainnat}